\documentclass[acmsmall]{acmart}

\usepackage{booktabs} % For formal tables

\usepackage[ruled]{algorithm2e} % For algorithms
\usepackage{graphicx}
\usepackage{epstopdf}
\usepackage{amssymb}

\usepackage{multirow}
\usepackage{longtable}
\usepackage{pdflscape}

\usepackage{color}
\usepackage{cancel}
\usepackage[normalem]{ulem}

\begin{document}
% Title portion. Note the short title for running heads 
\title[Video Skimming: Taxonomy and Comprehensive Survey]{Video Skimming: Taxonomy and Comprehensive Survey}  

\author{Vivekraj V K}
\orcid{0000-0002-4020-6703}
\email{vk@cs.iitr.ac.in}
\affiliation{%
  \institution{Indian Institute of Technology Roorkee}
  \department{Department of Computer Science and Engineering}
  \city{Roorkee}
  \state{Uttarakhand}
  \postcode{247667}
  \country{India}
}
\author{Debashis Sen}
\email{dsen@ece.iitkgp.ac.in}
\affiliation{%
  \institution{Indian Institute of Technology Kharagpur}
  \department{Department of Electronics and Electrical Communications Engineering}
\city{Kharagpur}
  \state{West Bengal}
  \postcode{721032}
  \country{India}
}
\author{Balasubramanian Raman}
\email{balarfcs@iitr.ac.in}
\affiliation{%
\institution{Indian Institute of Technology Roorkee}
\department{Department of Computer Science and Engineering}
\city{Roorkee}
  \state{Uttarakhand}
  \postcode{247667}
\country{India}
}

\renewcommand{\shortauthors}{Vivekraj V. K and Debashis Sen, et al.}

\begin{abstract}
Video skimming, also known as dynamic video summarization, generates a temporally abridged version of a given video. Skimming can be achieved by identifying significant components either in uni-modal or multi-modal features extracted from the video. Being dynamic in nature, video skimming, through temporal connectivity, allows better understanding of the video from its summary. Having this obvious advantage, recently, video skimming has drawn the focus of many researchers benefiting from the easy availability of the required computing resources. In this paper, we provide a comprehensive survey on video skimming focusing on the substantial amount of literature from the past decade. We present a taxonomy of video skimming approaches, and discuss their evolution highlighting key advances. We also provide a study on the components required for the evaluation of a video skimming performance.
\end{abstract}



\maketitle

\section{Introduction}
\label{sec:introduction}
Gigantic amounts of digital video are being produced for use in many areas such as education, entertainment, surveillance, information archival, etc. Due to the availability of large amounts of video data, there is an urgent need of corresponding tools and techniques for easy viewing, browsing and storing of videos. The sparsity of new information conveyed through hours of video data has motivated researchers to find ways to shorten videos to much smaller lengths so that their consumption is feasible, that is, a provision to comprehend  the video quickly in a shorter time is available. Video skimming is a process of generating a shorter version of the original video without spoiling the capability to comprehend the meaning of the whole video. 

Investigations on video summarization began about a quarter century ago with a focus on generating key frames (images) from the videos in order to provide a glimpse of the video contents. Although a sequence of images was shown on the screen, it was found insufficient for users to understand the video, especially in the case of long videos \cite{truong2007video}. However, the keyframe based techniques served the purpose of video browsing and indexing as well as for thumbnail representation of the video. Referring to such keyframe based summarization as static summarization, in recent literature, the focus has shifted to generating summaries as shorter videos by processing both the visual and audio content. This is called dynamic summarization, which improves the information conveyed by the summarization, as the generated shorter videos, called skims, consist of video segments and corresponding audio information. The key difference between static and dynamic summaries is the presence of motion  and audio information in the latter. Some of the key benefits of video skimming/ dynamic video summarization are: 
\begin{enumerate}
\item Conveying the plot of the video in shorter time.
 \item Reduction in transmission time for videos browsed over the Internet.
 \item Increases storage space utilization, by storing the video in its summarized form.
 \item Assimilation of information conveyed through multiple videos belonging to a topic (multi-video skimming; refer Section \ref{multiview}).
\end{enumerate}

\begin{figure}[t]
\centering
  \includegraphics[width=3.5in]{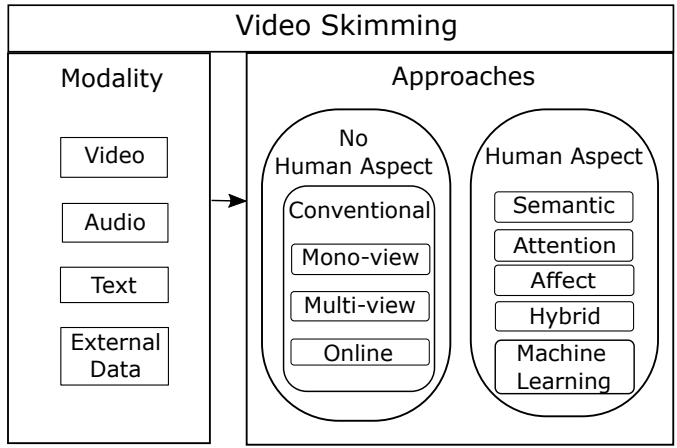}
  \caption{Taxonomy: Modality and human aspect based two-way $    $categorization.}
\label{fig:reviewclassification}
\end{figure}

However, there are a few key issues to deal with in video skimming. They are the processing paradigm, appropriate datasets for machine learning and evaluation, and evaluation strategy. Approaches under machine learning (ML) paradigm and those not using machine learning have been explored for video skimming. Conventional techniques not using machine learning work with handcrafted features, usually chosen by analyzing the video domain, and with decision-making strategies that are tuned to particular video domains. Decision making not tuned to a particular kind of data are generic in nature, but may not produce the best possible result. ML techniques learn patterns for decision making from annotated ground truth summaries available within the datasets. However, such techniques may not perform well on videos different from those in the datasets. This can be overcome by training on large datasets containing all possible kinds of videos from a domain, but generating such a large annotated datasets is challenging. Large datasets also pave the way for learning features \cite{panda2017weakly} along with decision making through supervised deep learning.  Further, unsupervised deep learning can also work through maximization of similarity between the video and its skim \cite{yang2015unsupervised,8099801}. Another crucial part of processing paradigm is the basic unit considered, which can be as small as a single frame to as large as a scene. This usually depends on the granularity of videos from a particular domain, that is, the smallest unit which can be meaningfully considered. However, smaller the basic unit, larger the number of computations required \cite{rasheed2005detection} and lessen the possibility that the temporal layout of motion \cite{8099801} is preserved in the video skim. Video skimming is a subjective task, which human beings can perform with ease. More often it is the human viewer who is the consumer of the summary, and as different viewers will have varying preferences \cite{del2018phd}, it is hard to find one summary that fits all viewers' needs. Therefore, it is imperative to include subjective criteria like informativeness, enjoyability, etc., into the video skimming evaluation process. We discuss the above key issues along with presently ensuing challenges and allied future scope in detail as listed in the organization of the article.

A few surveys related to video summarization are available in literature. Such a survey is given in \cite{money2008video}, which focuses on the modes of information (internal /external) used for generating summaries. Internal information refers to the visual, aural and textual content of the video, whereas external information is derived from the external world, such as viewer comments about the video, user ratings, review about the video on websites. A discussion about the attributes of video summarization techniques for static as well as dynamic summaries is done in \cite{truong2007video}. The paper provides an architectural view of different components of a video skimming system addressing both static and dynamic systems. The focus is on the variety of attributes that impact the way summarization is done, namely: video domain, video-skim unit selection, summary generation technique and the kind of summary. 
In \cite{Jiang2009}, the authors give in-depth analysis of video summarization focusing on the constituent temporal segmentation. In \cite{hu2011survey}, a brief description of  summarization techniques is given from the context of content based video indexing and retrieval. It categorizes the approaches of dynamic video summarization into redundancy removal, object or event detection and multi-modal integration. An overview of techniques for movie skimming system is discussed in \cite{li2006techniques}. They explain the structure of movies by highlighting the story units within a movie, such as broad scene, narrow scene, shot and frame. The paper organizes the approaches of movie skimming into a utility based or structure based classes. In utility based techniques, identification of salient objects and scenes is performed by building attention models. Structure based techniques perform hierarchical processing by utilizing the structure of the video: segments, shots and scenes. \cite{Oskouie2014} discusses semantic detection using a fusion of multi-modal features for soccer (sports video) from the perspective of video summarization and retrieval. The latest survey \cite{7750564}, discusses the specific challenges involved with egocentric video summarization. As none of the surveys in existing literature exclusively and broadly discusses video skimming, and the specific issues and components associated with it, a prime motivation lies for surveying dynamic video summarization. Many existing surveys have focused either on summarization in general, with emphasis on static summarization, or directed towards a specific video domain.  

In this article, we provide a comprehensive survey of video skimming approaches. We present a taxonomy of the approaches and discuss different paradigms considered to implement various essential parts of a skimming system. Our taxonomy is shown in Figure~\ref{fig:reviewclassification} (elaborated in Section~\ref{Taxo}) reflects how the field has evolved, especially in the last decade, with the primary categorization being based on whether aspects of human understanding of videos are considered or not. Attention, affectiveness and semantics are key aspects through which humans understand the world and these aspects have been used considerably for video skimming. Whether human aspects are considered or not, a video skimming technique aims to provide a summary that would score high on completeness, coverage, concise and context \cite{ren2010video}.

The survey presented in this paper is one of the first of its kind focusing only on dynamic summarization of videos. With the recent surge in video skimming research, enough material has become available in the literature to provide such a survey that not only discusses the approaches available but also their evolution highlighting key advances. In our survey, the discussion involves seminal approaches under each classification and focuses on the evolution of approaches. The article also provides an in depth discussion of the different video skimming evaluation techniques (refer Section \ref{sec:evaluation}) that have been developed over the past decade. We cover from subjective evaluation, where we discuss evaluation based on criteria such as interestingness, informativeness, enjoyability, etc., to the use of objective evaluation done in contemporary datasets with the help of ground truth. 

The organization of the article is as follows. We begin in Section~\ref{concepts} by presenting a generic structure of a typical video skimming system, the video domains involved in video summarization and the taxonomy considered in surveying the literature. A detailed exploration of each of the branches in the taxonomy is considered in Section~\ref{sec:techniques}. Different datasets that are available for video summarization along with the evaluation methodologies are discussed in Section \ref{sec:evaluation}. In Section~\ref{subs:scope}, present challenges and future direction are discussed followed by concluding remarks.

\section{Survey Preliminaries}
\label{concepts}

\subsection{Terminology used}
We enlist a few terms that we shall use for ease of reference:

\begin{itemize}
\item Video-skim-  shortened version (summary) of the original video.

\item Skim ratio/ summary length-  the length of skimmed video expressed as percentage of the length of the original video.

\item Skim unit-  basic block of the video considered while processing. These blocks can be shots, scenes (structured video), events/segment (unstructured video) or just a sequence of frames.

\item Skim unit length- length of each skim unit.

\item Structured video-  videos which mostly follow a script, like movies, news, etc. Others are referred to as unstructured videos like consumer videos, personal videos/ user videos/ home videos, surveillance video (refer Table \ref{videodomains}).

\end{itemize}

\begin{center}

\tiny
\begin{table}[t]
\centering
\caption{Different video domain and their summarization criteria}
\begin{tabular}{|c|p{.30\textwidth}|p{.30\textwidth}|p{.10\textwidth}|}
\hline
\textbf{Video Domain} & \multicolumn{1}{c|}{\textbf{Description}} & \multicolumn{1}{c|}{\textbf{Summarization Criteria}} & \multicolumn{1}{c|}{\textbf{References}} \\ \hline
\textit{News} & Structured video consisting of news reader/anchor and their interaction with correspondent. Interview/ discussion among a group of people. Often interleaved with commercial advertisements. & Selection of prominent sections of the news like headlines, stories, critical points in interview/ panel discussion. & \cite{757478,taskiran2006automated} \\ \hline
\textit{Sports} & Consists of sports activity recorded by multiple cameras. Typically consists of play, replay and break sequences. 'play'- where the game is going on, 'break'- pause time during the game. &  Event identification, event corresponds to the most attractive part of the sport e.g. scoring a goal in soccer/ football. These moments are usually called `hotspots'. & \cite{xu2008novel,yoshitaka2012personalized,5133759,5582561} 
\cite{6422365} \\ \hline
\textit{Rushes} & Unedited recording of T.V programs, which contain junk frames, clapper board and duplicate segments due to re-recording of the same activities. & Selection of unique parts of the video by removing all unwanted and duplicate recordings. & \cite{5993544,Valdes:2008:BTB:1463563.1463588,4907069}  \\ \hline
\textit{Movies} & Highly structured sequence of shots and scenes. In addition may contain text information in the form of subtitle file. & Identification of significant parts necessary for story narration. & \cite{dong2010video,dong2014iteratively,evangelopoulos2013multimodal, ren2010semantic,tsai2013scene,Chen2003358,PFEIFFER1996345,vasconcelos1998bayesian,kannan2015you,7362797} \\ \hline
\textit{Surveillance} & Continuous recording of activity using either a static or moving camera. Single or multiple cameras & Event detection and anomaly identification. & \cite{6838965,fu2010multi,7121011,7547309} \\ \hline
\textit{User/Consumer} & User recorded videos having significant uncorrelated camera motion, usually of few minutes. &  Identifying interesting parts of the video. & \cite{Wang:2014:RSU:2647868.2655013,gygli2014creating,cong2012towards,potapov2014category,hong2009event,hong2011beyond,wang2012event,Zhang:2012:MSS:2155555.2155565,peng2009user}
 \\ \hline
\textit{Egocentric} & Recording of day long activity of the person using head/chest mounted camera. & Important parts of the video conveying the typical activities of daily living of the camera wearer.  & \cite{xu2015gaze,lu2013story,yaohighlight,Varini:2015:EVS:2733373.2806367,ho2018summarizing} \\ \hline
\end{tabular}
\label{videodomains}
\end{table}

\end{center}

\subsection{Taxonomy}
\label{Taxo} 
The taxonomy considered for surveying the video skimming literature is shown in 
Figure \ref{fig:reviewclassification}. A two-way categorization is done, where one is based on the data modalities employed and the other is based on whether human aspects are considered or not. The data modalities represent an overview of the variety of features used for identifying significant video skim units. Apart from video and audio, text modality refers to the subtitles of the video either available separately or extracted from the audio using Automatic Speech Recognition (ASR). External data refer to additional information (like video tags, comments, viewer behavior in terms of eye gaze, etc.) used either independently or together with video, audio and text.

The approaches which do not consider human aspects are called conventional approaches. These are based on statistical processing of low level features such as color, intensity, orientation, texture, etc., from videos. Preliminary work under conventional approaches primarily focused on single video summarization. With the emergence of video surveillance and video search services, video skimming has been expanded to multi-view/ multiple videos. Further, with the advancement of embedded systems (wireless video sensor) and video streaming services, where the entire video is not available for processing, online video skimming has been invented. Thus, conventional techniques are sub-categorized into mono-view, multi-view and online. 

The approaches employing human aspects create summaries based on well-known models of human understanding of videos.  We draw motivation for having such a category from \cite{frintrop2010computational}, wherein emphasis is given to the computational modeling of human attention phenomena. The techniques discussed under this category, measure some of the human influencing parameters like objects, object  category, salient objects, emotional impact on the viewer etc., to determine the suitability of video segments to be included in the skim. The approaches which consider human aspects are further classified into those considering human attention, affectiveness of contents, media semantics and learning from examples, as these are the major categories which have been effective in video content summarization during the last decade. Semantics, attention and affect based approaches are usually based upon what video parts humans might consider important and how they understand the video through these important parts. The hybrid category refers to any combination of attention, affect and semantics in order to have a more sophisticated human aspects model for performing video skimming. Machine learning based approaches create a model from given human created summaries using which video skimming is performed. Although human aspect based skimming can be categorized again into mono-view, multi-view and online techniques, the multi-view and online cases for such approaches have not been explored substantially in literature, and hence we do not have such a categorization.

\subsection{Generic Video Skimming System}
\label{genericblockdiagram}

\begin{figure}[!h]
\centering
  \includegraphics[width=9cm]{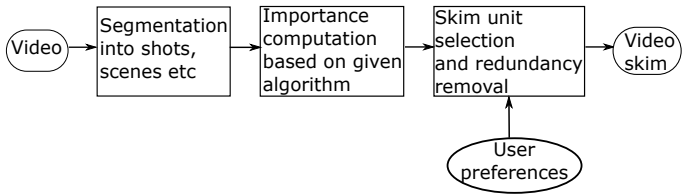}
  \caption{Generic block diagram of a video skimming system. }
\label{fig:blockdiagram}
\end{figure}

The block diagram showing the essential components of a typical video skimming system is given in Figure \ref{fig:blockdiagram}. 

\subsubsection{Segmentation: }The segmentation (pre-processing) block deals with the temporal segmentation of the given video into smaller units called skim units. Usually, this block segments a video into smallest comprehensible parts and these units are processed independently. A smallest comprehensible part refers to a set of minimum number of frames that contains activities collectively conveying some meaning. 

Segmentation was initially performed by making use of color histogram difference \cite{chen2009novel, gao2009dynamic} or intensity difference \cite{chu2015video} that imply a change in the visual contents, which works well for structured videos. Novel segmentation techniques are proposed for user and egocentric videos, such as motion based segmentation \cite{gygli2014creating} and change point detection \cite{potapov2014category}. In \cite{xiang2011affect}, SVD of color histograms followed by clustering is used for segmentation. Scene identification is performed by clustering the shots using color histogram in \cite{Chen2003358}, whereas graph partitioning based technique is used in \cite{ngo2005video}, \cite{tsai2013scene}, where in the shots are represented by weighted undirected shot similarity graph. The nodes of the graph are identified with the shots and the edge weights signify the similarity between the shots measured using color and motion. External data are also used for segmentation, such as in \cite{xu2015gaze}, which uses eye gaze information. Audio based segmentation \cite{li2011static}, \cite{5133759} has also been considered for videos having mostly static visuals.
 
Segmentation can help in reducing the computation time as a representative frame \cite{rasheed2005detection} can be used instead of processing all the frames in the segments. A detailed survey of different video segmentation techniques is given in \cite{SMEATON2010411, koprinska2001temporal}. In structured videos the video segments are well-defined shots and scenes, which can be easily detected; use of these segments as skim units will help in maintaining temporal connectivity within skim units. Unstructured videos do not have well-defined boundaries, and usually, it is a matter of choice to choose video segments. During this choice, the granularity of video segments becomes an important parameter that directly impacts the computational time required for feature extraction and processing. For example, it is suggested in \cite{yeung2014videoset,Sharghi2016} to use 5 to 10 second video-skim unit for processing egocentric videos. Online processing can be performed frame-wise or by considering a set of frames as all the video data is not available at once.

\subsubsection{Importance computation: }
The importance computation block calculates the importance of skim units. This importance usually depends on multiple low and high level features, such as motion, color, aesthetics, semantics, etc., extracted from the units. Video skimming can be performed by computing the skim unit importance either directly or indirectly. In the direct method, importance is computed considering all the frames in an entire skim unit, thus making the computation expensive. Whereas, in the indirect method, importance of a skim unit is computed on key frames extracted from it, thus making the computation faster. The fundamental approach to summarization is identification of important video skim units. This importance is manifested in terms of a score, by considering the quantity of chosen features in conventional techniques. In attention based techniques,  it is the amount of attention grabbing contents in the skim unit; in semantic based techniques, it is the richness of semantics measured; in affect based techniques, it is the magnitude of impact on the viewer's emotions; while the learning based techniques reduce the error between the predicted and expected values. The specific methods employed will be discussed in Section~\ref{sec:techniques}.

\subsubsection{User preferences: }
The user input block accepts user requirements for the skimming to be performed. It typically includes skim length, kind of skimming to be performed (overview or highlights) and any other parameter which is customized for an application scenario. Highlights give preference to representation of significant events of the video by the skim, usually relevant for sports and surveillance video skimming, and overview gives preference to representation of the entire video content by the skim, usually relevant for movies. 

The user preferences help in personalizing the skim as per the viewers' taste. As in for movie skimming, some viewers may be interested in watching specific kind of scenes like action, romance, comedy, etc. The personalization can also be in terms of specific features of the video like people, objects, events etc. \cite{kannan2015you}, \cite{xiang2011affect} or it can be in terms of viewing time. The preference can also be given in terms of the kind of frames~\cite{darabi2015personalized}, \cite{han2011personalized} the viewer likes to see in the skim. Personalization can be done non-intrusively by observing viewer behavior \cite{yoshitaka2012personalized} or by learning from viewer comments \cite{chung2014personalized}.

\subsubsection{Skim unit selection: }
The skim unit selection and redundancy removal block decides which skim units should be included in the video skim based on skim unit importance, skim length and other user parameters. This block also removes similar skim units within the video skim in order to achieve the best possible video skim which covers the required details in the original video.

A typical skim formation strategy can be categorized into score maximizing and redundancy removal methods. In scoring based methods, the video skim units are given a score in terms of the importance criteria used, followed by selecting most important segments. In redundancy removal, relative comparison among the video skim units is done so that unique set of segments are selected into the summary. Most techniques are based on scoring, while a few like \cite{4907069}, \cite{gao2009dynamic} focus on redundancy removal. Both \cite{4907069}, \cite{gao2009dynamic} perform clustering to identify representative shots from each cluster. A graph representation is used in \cite{Chen2003358} for identifying the prominent scenes from the video. Visual similarity between the shots is put to use in \cite{laganiere2008video} for duplicate removal. \cite{zhao2014quasi} tries to remove duplicate segments by comparing with a dictionary of essential segments.

\begin{figure}[t]
\centering
\includegraphics[width=\textwidth,height=6cm]{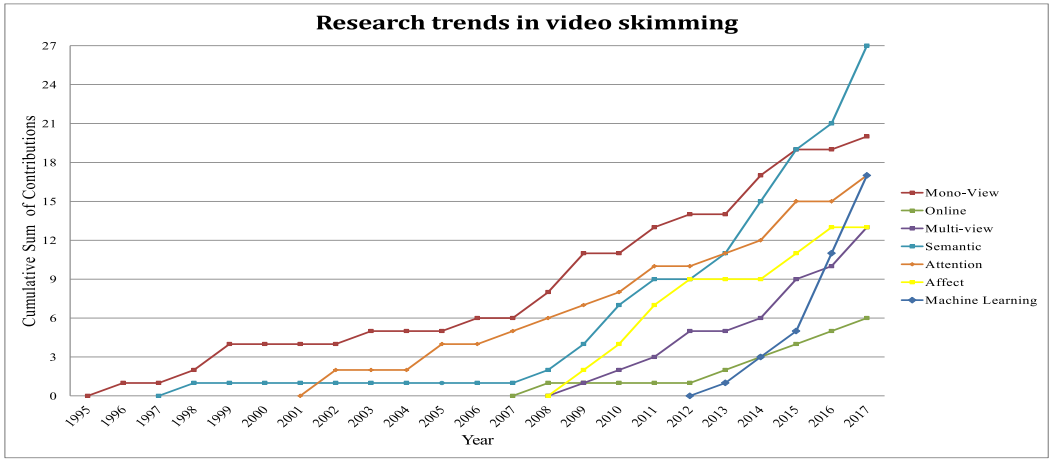}
\caption{Research trends in video skimming. The graph is constructed by considering all the publication considered in this survey under each category of the taxonomy. Emphasis has been given  to consider most publications in reputed journals and conferences\protect\footnotemark over the past decade as the last generic survey on video summarization \cite{truong2007video} appeared in the year 2007. (Best viewed in color)}
\label{fig:growthvs}
\end{figure}

\footnotetext{Conferences and Journals of Relevant Reputed Societies like ACM, IEEE Computer society, IEEE Signal Processing Society, IEEE Circuits and Systems Society, IEEE Communications Society, and other Journals whose current (2017) JCR $\geq$ 1.}

\subsection{Growth of Video Skimming}
The growth of video skimming techniques in terms of the taxonomy considered is shown in Figure \ref{fig:growthvs}. The graph shows yearly cumulative sum of all the contributions (considered in this survey), along with the nodal papers that appeared in each of the categories as shown in the taxonomy. The graph shows that the diversification of video skimming approaches has started happening only in the last decade, emphasizing the importance of this survey and the taxonomy presented. It would be valuable to look into different video domains on which video skimming has been performed, Table \ref{videodomains} gives a brief description about different video domains and related typical criteria used to generate video skim.

\section{Systematic Overview of Techniques}
\label{sec:techniques}
In this section, we provide an overview of video skimming approaches in the literature following the categorization from the taxonomy shown in Figure \ref{fig:reviewclassification}. In addition to video skimming certain keyframe techniques like \cite{cong2012towards,dang2014heterogeneity,Mei:2013:NSV:2487268.2487269,ji2019query,7780487} are also included for their representative contribution which can be extended to video skimming. The work in each category of the taxonomy is sub-grouped using suitable criteria like video domain, skimming parameter, etc., providing a structure that would simplify locating required information. Within each sub-group, the works are ordered chronologically to elicit the development in reference to existing ones. Further, each sub-section begins with the major idea of the category along with a broad relational analysis among the works which acts as an index for easy browsing and comprehension of the survey.

\subsection{No Human Aspect (Conventional)}
The conventional approaches deal with extracting a variety of features primarily from video modality and determining the importance of skim units for inclusion in the video summary. The low level features include edge histogram, wavelet features, intensity, etc. The middle level features include motion, face, person, aesthetics, etc., and high level features like follow-object, event detection, etc. The features are chosen in accordance with the video domain (see Table \ref{videodomains}) considered and the type of summary (overview/ highlight) to be formed. Audio features like energy, pitch, zero crossing, etc., are used and audio is also used for obtaining the speech transcript.

The techniques have evolved by addressing the varied challenges posed by different video categories, starting from structured videos like movies and TV episodes to the contemporary unstructured ones like user and egocentric videos. Following a similar pattern we have sub-grouped the works under each sub-section on the basis of video domain. In Section \ref{monoview}, we discuss the mono view video summarization followed by multi-view in Section \ref{multiview}, online in Section \ref{online} and discussion in Section \ref{NonHumanDiscuss}.

\subsubsection{Mono-view/ single video: } \label{monoview} The works discussed in this section are applied on single videos and address the specific challenges of the video domains. Movie and news videos are scripted and contain significant audio information. Therefore, audio information analysis and graph-based modeling of scenes can be experimented in such videos \cite{Chen2003358}. Rushes videos are also scripted TV episodes, but they are unedited videos which contain significant repeated segments and junk frames. These videos can be processed by solving each of these problems sequentially \cite{4907069}, or performing segment categorization and using audio cues to identify the segments which are worth preserving in the summary \cite{5993544}. Sports and surveillance videos have a long duration of uninteresting content, where highlight detection can be done using the motion and audio cues \cite{5133759}. User and egocentric videos are unscripted which may contain significant motion and redundant segments. Redundancy can be removed by measuring video activity level and processing the video in terms of shots and scenes \cite{laganiere2008video,gao2009dynamic}. Representative segments from videos can be identified through sparse coding \cite{cong2012towards}, and optimization techniques can be used for assigning the summary duration among the video segments by adapting the playback speed of the video \cite{6671471} and reducing object space dimension \cite{6247852}.

{\bf Movie and News Videos. } The automatic movie trailer creation approach of \cite{PFEIFFER1996345} is one of the earliest in this category. The authors highlight the requirements of a good abstract, which are based on physiological evidence from perception to propose a video skimming system. The process  involves video segmentation followed by selecting high contrast and action scenes to form the video trailer. Whereas in \cite{Chen2003358} the trailers are generated by first performing shot clustering, as a collection of shots might deliver a summary that allows better understanding of the video. Segmentation is performed by frame-wise color histogram analysis. The temporal interconnections between various shots is modeled using a scene transition graph, where each node represents a scene and directed edges show the temporal order between the shots in different scenes. Summarization is based on analyzing the graph to find interesting scenes like the ones among which more edges are present, and those with many small shots implying significant camera movement.

Possibly the earliest approach to consider integration of different modalities for video summarization is presented in \cite{smith1998video}. In the paper, scene segmentation is performed using inter-frame color histogram difference. Considering text as important, text detection is performed to segment frames containing textual information by making use of the fact that textual regions have high contrast with the background. Moreover, assuming the presence of face signifies important activity, face detection is also considered. Frequency-inverse document frequency is used to detect key words from speech, considering the availability of the speech transcript of the video. Audio skimming is performed by considering the key phrases obtained that begin with the key words and their importance. Apart from face and text presence, scene similarity, object and camera motion are also considered. A final video skim is generated using visual and audio based importance curves. In~\cite{nam1999dynamic}, an adaptive nonlinear sampling technique is used for generating video skims. The sampling rate is determined by discretizing the motion activity curve and mapping them to predefined quantized values. Action events in the video are identified by measuring the amount of red pixels for blood and flame. Audio information is used to classify the scenes into emotion or no-emotion classes. Based on the given summary length, first skim units with important events (violent and emotion) are selected, then other skim units are selected uniformly from the video for remaining summary length. A skimming system for news is presented in \cite{757478}, where hierarchical partitioning of the video is performed for detecting specific contents (commercials, anchor persons) using multi-modal features. Commercials are detected using audio cues, anchor persons are segmented using Gaussian mixture models and text processing is done to identify story units. The system is tested for facilitating browsing of news video. 
 
\cite{taskiran2006automated} performs summarization of documentaries, presentations and educational videos by extensively making use of audio information through audio analysis, automatic speech analysis and recognition.  Here only audio information is used as the visuals do not change much in the above videos. A proprietary toolkit called `cue video toolkit' \cite{Ponceleon:1999:CAM:319878.319937} is used for shot boundary detection, speech recognition and indexing. Audio phrase boundary and pause duration between sentences is used for video segmentation. Segment importance weights are calculated using word frequency within each segment. Dominant word pairs are detected for giving additional increments to the segment weights, as some specific phrases may have special significance. The video skim is generated by selecting the segments having highest weights until the given skim ratio is satisfied.

In \cite{li2011static}, dynamic summaries are obtained using both video and audio information. ASR is used to obtain speech transcripts from audio. This textual information is used to segment the video based on n-grams. A modified Maximal Marginal Relevance (MMR) \cite{carbonell1998use} based technique is used to determine the importance of segments. MMR is based on the information content of the segments.

{\bf Rushes Videos. } Summarization of rushes video (refer Table \ref{videodomains}) is done in \cite{4907069}. Video segmentation is performed by measuring the activity level of each frame, where activity is determined by frame wise difference of the color components. The cumulative activity from all the frame in a shot determine its score. Junk frames are removed by histogram analysis and clapper board frames are removed using luminance energy. Adaptive clustering is then used to filter out retake shots. Key frames are extracted from each active shot one each at the beginning, middle and end of the shot. Similarity between the keyframes based  on histogram correlation and image difference is used for shot clustering. Representative shots are selected from each cluster using their length and activity level. In \cite{5993544}, a hierarchical Hidden Markov Model (HMM) is used for summarizing rushes videos. The first layer of HMM categorizes the video segments into stock, outtake, shaky and second layer identifies the camera motion in each of the shots. `Stock' is a shot that is worth keeping, `outtake' is a shot which will be removed during editing and `shaky' is a shot which will be definitely discarded. Clapper board frames are discarded by identifying near duplicate key frames \cite{Ngo:2006:FTN:1180639.1180827, 4276722}. Valid shots are those recorded between the `action' and `cut' commands of the director and these keywords are recognized from the audio signal. Retake shots are avoided by selecting the last recorded shot of a particular dialog. Shots are selected to be included in the summary based on pairwise comparison with other shots in terms of representability score derived using object, visual event and audio events in the shot. 

{\bf Sports and Surveillance Videos. } 
Identifying interesting events in a soccer game using audio features such as block energy and repetition index is experimented in \cite{5133759}. Block energy is the energy of a set of audio frames. Repetition index measures the repetition of words like `goal' in the audio segments. 

Constraint Satisfaction Programming (CSP) is used in \cite{7368910} to specify a set of conditions for summary generation. The authors suggest a clear advantage of summary generation rules (specifying conditions) over summary generation algorithms, as new constraints can be added/removed at any time without any alteration. Unlike previous techniques which performed video segmentation, here for each required feature, segments are identified. The features are represented as a set of segments where the required feature exists. A segment is identified using the starting frame and number of frames (duration). The CSP is implemented using Choco CSP solver \cite{jussien:hal-00483090}, with the constraints specified in terms of  summary length and description of segments (e.g. Containing applause, presence of face, etc.) to be preserved in the summary.

{\bf User Videos. } 
In \cite{laganiere2008video}, spatio-temporal features are used to measure the activity level in a video frame. The interest points for activity level determination in a video sequence are pixels which show large variations in intensities in both spatial and temporal dimensions. The activity level is used to identify the key frames in the video and segments are selected around the key frames by using a suitable segment length based on the activity level of the surrounding frames. Distances between the segments are computed based on color histogram, to reject those which are similar to others from being part of the summary. The final summary is formed by incrementally adding the clips based on the decreasing level of their activity until the user specified summary duration is met. In~\cite{gao2009dynamic}, video skimming is achieved by removing redundancy hierarchically, first at frame level and then at the segment level. Key frames are identified using leader-follower clustering algorithm \cite{rasheed2003scene}. To get the segments or skim units, shots are first determined through video segmentation using color histogram and optical flow motion features. Then scene detection is done by using the normalized cut approach to cluster the shots, and the scenes obtained are considered as the segments. The segment level redundancy is removed using Smith Waterman algorithm \cite{smith1981identification}. Redundancy removal at segment level is justified by suggesting that two similar segments with large camera movement or long shot duration will have different key frames.  Frame level redundancy is removed using hierarchical agglomerative clustering. Video summary is generated by first calculating the scene importance based on the shots contained and then by calculating key frame importance based on motion, presence of face and audio energy. 
The work in~\cite{peng2008aesthetics} proposes to generate a music-video  style of video summary given a home video and background music. Important content determination through video analysis and music chunk, tempo determination through audio analysis are considered. The technique starts with video segmentation performed through histogram based shot change detection \cite{hanjalic2002shot}. Shot lengths are estimated based on music chunk by aligning shot boundaries with music onset using a suitable function. Trimming of the segmented shots is done to fit them into the estimated shot lengths by extracting visually important sub-shots. Transition between sub-shots is decided based on average music tempo of the shot. The summary is obtained as a collection of most important sub-shots.

\cite{cong2012towards} discusses sparse dictionary selection technique for summarizing consumer videos. As the summarization process is similar to sparse selection of segments from the original video, the same is performed here using the procedure of \cite{yuan2006model}. Feature selection is done by extracting features from all frames in the video using the dictionary selection algorithm of \cite{yuan2006model}. A weight curve on video frames is formed based on which key frames are identified. Skimming is done by selecting a set of frames around the keyframe as per a pre-defined minimum segment length. 
 In \cite{gygli2014creating} interestingness of segments is measured using aesthetic quality, faces and objects. `Superframe'- a motion based sub-shot segmentation method for user videos is also proposed. Features are combined using a linear model for giving an interestingness score to the segments.

{\bf Generic Methods. } The works in \cite{6671471,gygli2015video,6247852} define an objective function for each of the criteria that the summary needs to satisfy and an optimization framework will derive the summaries. \cite{6671471} proposes a resource allocation framework, where summary time is distributed among various segments by adapting the playback speed. Video segmentation is performed utilizing the cues from video production \cite{owens2015television}. Each segment is assigned a base benefit based on the visual content (still and dynamic). A discrete set of playback speed are assigned to each of the segment followed by Lagrangian relaxation and Convex-hull optimization \cite{everett1963generalized} methods to maximize the benefits (base and user preferences).

In \cite{dang2014heterogeneity}, a new image feature is given and called the heterogeneity image patch (HIP) index, which is an entropy based technique to measure the level of non-redundancies among the image blocks. Each video frame is divided into fixed size non-overlapping  blocks /patches and correlation  between the patches  of the frame is found out by using  a threshold  and distance metric (sum of absolute differences).  An entropy computation considering the correlations gives the HIP index for a given image/ video frame. The HIP based dissimilarity between the original video and a video skim is then considered to get an optimal video skim minimizing the dissimilarity.

\cite{del2017active} proposes an interactive customization of video skimming process by asking questions to the viewer. A probabilistic approach based on active inference in conditional random fields (CRF) \cite{6751398} is used to determine the segments to be included in the summary. Motion and illumination changes are used to derive segments from the video (around 2.5 seconds). Initially the viewer is shown the most prominent objects and places in the video followed by two question: 1) whether the viewer want to see this segment in the final summary? 2) should similar segments be included in the summary? Depending on the viewers response, an energy function of CRF (that includes terms for selecting static segments and a term that selects diverse segments) used to determine the next segment to be shown to the viewer. This iterative sequence continues till the viewer is satisfied with the generated summary.

\subsubsection{Multi-view / multiple videos: }
\label{multiview}
In multi-view videos, a scene is captured with different viewpoints (multiple cameras); unlike single view video, the challenge here is to identify duplicate skim units across videos. While near duplicate keyframe identification on the basis of visual similarity is performed in \cite{hong2011beyond}, dimensionality reduction based approach is suggested in \cite{7532345} for user videos. Typically sports and surveillance videos are multi-view by default. A graph based duplicate detection is performed in \cite{fu2010multi,7121011} for such videos, whereas a two phase redundancy removal is performed in \cite{6838965} for intra-view and inter-view video segments. Generic methods  \cite{chu2015video,8099938,6909934,7952384} in this category utilize external information like similar videos or photos for determining the segments from specific view to be preserved in the summary.

{\bf{ User Videos. }}
The authors of \cite{hong2009event} and \cite{hong2011beyond} proposed a technique for easy browsing of videos retrieved through a user query. Suggesting that the videos retrieved based on a user query will have overlapping content, an automatic video summarization of the multiple videos belonging to a category is proposed. This is performed by identifying key shots based on a similarity threshold on keyframes and then ranking the shots according to their informativeness scores. Shots are grouped based on visual similarity \cite{lowe2004distinctive} into Near Duplicate Keyframe (NDK) groups, and a representative shot from each group is extracted.  The informativeness scores are determined by the relevance of a video to the given query and normalized significance of its representative shots. Key shots threading is performed to reduce the time lag between chronologically ordered shots for the skim.  Shots are selected to form a video skim by optimizing both the sum of scores and time lag. This work is improved in \cite{wang2012event} by reducing the computational load due to the NDK grouping. Instead, they use metadata such as tags for selecting relevant shots. In \cite{ji2019query}, the videos are segmented into shots \cite{yuan2007formal}, and the middle frame is considered as candidate frame. Images relevant to the considered topic of the set of videos are used to guide the selection process. The candidate frames are considered as basis vectors for sparse coding all the frames i.e., candidate and related images. Coefficients are determined for each candidate frames by employing them in a sparse coding framework to jointly reconstruct all candidate frames and images. These coefficients act as important scores that signify which frames/shots are to be part of the summary.

An objective function based mapping is done in \cite{7532345} for summarizing multi-view videos. The mapping is performed on the frame proximities from D dimensional space to d dimensional space, where d $\ll$ D. In case of multi-view videos, there are intra-view and inter-view proximities between the video segments. Intra-view proximity is measured using a Gaussian kernel while inter-view proximities are found using Scott and Longuet-Higgins algorithm \cite{scott1991algorithm}. Both the proximities have been accurately mapped to a lower dimensional space using an objective function. Finally a subset of frames are selected based on their ability to reconstruct all other frames of multi-view video.

\cite{Zhang:2012:MSS:2155555.2155565} summarizes multiple videos tagged with Geo-spatial metadata to provide a route plan for sightseeing. 
Geo-spatial meta-data is captured using Global Positioning System (GPS) and compass, available in most smart-phones. Videos are divided into sub-shots using GPS location and camera movement. Selection of sub-shots covering a particular location (known prior) is done with the help of location co-ordinates captured by GPS and field of view of the camera. Video summary is formed by assembling significant sub-shots. Several such summaries are combined to provide a travel plan to the users.

{\bf {Sports and Surveillance Videos. }}
\cite{fu2010multi} discusses a method  for summarizing multi-view videos, mainly that of sports and surveillance. The videos are first segmented into shots using an activity based video representation of \cite{xiang2004activity}. Shot importance is computed using a Gaussian entropy fusion model that employs color histogram, edge histogram and wavelet features. Another importance score is determined by the presence of face in the video shot. Shot similarity is measured using spatial, and temporal measures and shots across different views are compared through a hyper-graph representation.  An edge in a hyper-graph called a hyper-edge, connects a set of shots across different views. The hyper-graph based representation is used to avoid fusion of similarity values as a fused value will not be able to represent the individual relationships between shots in terms of different measures. The hyper-graph is then converted to a spatio-temporal graph using \cite{sun2008hypergraph}, in which the node contains the shot importance score and edge weights determine the similarity between shots. Event based shot clustering is performed using random walks followed by multi-objective optimization to generate the video skim. In \cite{7121011}, instead of modeling all videos using a hyper-graph, pairwise comparison is performed by representing the videos as bipartite graph and mapping the similar shots using Maximum Cardinality and Minimum Weight (MCMW) \cite{shokoufandeh1999applications}.

A wireless video sensor (WVS) is a battery powered device with low power requirement. ``It consists of a server and several wireless video sensors, which stream the recorded videos to the server'' \cite{6838965}. Video streaming is known to have high data rate, and this implies large transmission power requirement.  As the WVS has low power design, embedding summarization algorithm into the sensor is explored in \cite{6838965} to reduce the amount of data transferred to the server. 
 A low complexity multi-view summarization system is implemented. The system is implemented in two parts: intra-view  and inter-view stages. In intra-view stage, similar contents are grouped by an online clustering algorithm using the MPEG-7 color layout descriptor \cite{959135} of each frame. Based on the parameters of the clusters, frames from prominent clusters are selected to form the single view summary. Inter-view summarization is performed using view selection \cite{6103271}, where a view among those captured by different sensors at the same time is selected by thresholding on the foreground color layout feature and size of the foreground. It is assumed that the sensors have broadcast facility to transmit the feature to nearby sensors. Finally the selected frames are transmitted to the server.

{\bf{Generic Methods. }}
The techniques  \cite{chu2015video,8099938,6909934,7952384} utilize the co-similarity or the co-information from similar videos or photos to perform effective summarization. These methods can be applied for both single video and multiple video summarization. In \cite{chu2015video}, video co-summarization is suggested, which is summary generation by considering a collection of videos on a particular topic. First, video segmentation is performed on each video by thresholding on the sum of squared pixel-wise difference between consecutive frames. Segments from all videos are represented as nodes in a graph and visual co-occurrence is estimated by extracting complete bipartite sub-graphs. The summary is generated by including segments that occur frequently across videos, by selecting the sub-graphs with most nodes (most frequent visual co-occurrence). \cite{6909934} jointly summarizes photos and videos related to a common topic by mutually relating them in a beneficial way. A K-NN graph between the photo streams and videos is built to group relevant photos and videos, helping in reducing the diversity among photos and videos. The video frames and corresponding matching photos are the nodes of the graph; the similarity between the video frames, and the similarity between videos and photos is considered as edge weight. Video summarization is performed by applying diversity ranking algorithm \cite{6126239} on the above graph, which tries to find a subset of frames based on an optimization function.

\subsubsection{Online: } \label{online}
Online techniques do away with the need of complete video being available for summarization task. These techniques work on the available parts of the video, for example, as in live video streaming. Locally optimal representative segments are selected and shown as a summary. Rushes videos are summarized in an online fashion in \cite{Valdes:2008:BTB:1463563.1463588}, by constructing a decision tree and evaluating the cumulative frame importance over a fixed time period. On the other hand, generic proposals in this category involve creating an online dictionary \cite{zhao2014quasi,7532342} to discriminate among redundant and unique segments. Optimization based techniques are suggested in \cite{8099680} for identifying the representative segments in an iterative way.

{\bf {Rushes Videos. }}
\cite{Valdes:2008:BTB:1463563.1463588} performs online summarization of rushes video. A dynamic binary decision tree is constructed, where for each incoming segment two nodes are created and appended to all the leaf nodes of the tree, one node represents the decision of `selection' and other represents `not selected'. A time window limits the number of segments that will be considered in deciding, whether the segment represented by the root node of the tree should be included or excluded from the summary. Each path from the root node to leaf nodes is evaluated in terms of the sum of importance score achievable from all the nodes in the path. The importance of each segment is determined based on: number of frames, similarity between segments and activity level. The path that maximizes the sum of scores helps in deciding the label for the root segment.

{\bf {Generic Methods. }}
\cite{almeida2013online} presents a technique for online summarization by processing the video in the compressed form. A frame is reduced to an image which is about 1/64 of the original size and represented by its color histogram. Grouping of video frames is performed using ZNCC \cite{martin1995comparison} as a distance metric. Selection of group of frames for a summary is done based on setting a threshold for the distance. A real time video summarization on mobile devices is suggested in \cite{7532342}. Video is segmented with the help of motion as in \cite{gygli2014creating}. Interestingness of a segment is derived from a linear combination of interestingness measures obtained using colorfulness, global camera motion and dictionary created using color histogram. The most interesting segments are included in the summary.

In  \cite{zhao2014quasi}, video summary is generated by eliminating redundant segments over time by comparing with a dictionary of segments created online. Segments are selected to be a part of the summary if reconstruction error to generate it using the dictionary is greater than a certain threshold. Whereas in \cite{8099680}, dictionary creation is performed using optimization techniques. Since the optimization techniques are not effective in an online scenario (as complete data is not available), an incremental way of selecting the representatives is shown with theoretical proofs and experiments on video summarization. Video is processed in batches of segments, previously selected representatives and the new batch of data is used to update the representatives. A dissimilarity based sparse subset selection \cite{7364258} is adopted in this work for incremental subset selection.

\subsubsection{Discussions: }
\label{NonHumanDiscuss}
The conventional techniques depend either on pixel level features or determine segment level/ frame level importance for deciding on the significance of the video segment. The major task is that of identifying the relevant set of features to be used based on the video domain. Selection criteria (refer Table \ref{videodomains}) for video summarization depending on the video domain is also done by defining objective functions. Only visual features are used extensively with very few techniques trying to utilize the available aural or text features. Since a video is essentially a combination of multiple modalities, a kind of fusion mechanism should be used to give appropriate preference to each of the modalities.

For multi-view videos, graph based modeling is found to be useful for representing the inter-relationship between views. Here, the summarization is divided into inter-view and intra-view phases. This two-phase processing is essential for removing redundancy within the view and across views. Online approaches thrive on extracting simple features that can be retrieved quickly from the available frames of the video based on which decision for summary generation is made. Whereas real time video summarization techniques \cite{7532342} are the ones which consider the amount of time spent in generating the video summary. Not much focus has been given for processing time till now, but inevitably will be considered as an important factor in future.

These techniques do not consider identifying the user level or high level abstractions based on which a human viewer will decide the segment importance. As the techniques rely on extracting features there is limited scope for personalization to viewers' expectations. The purpose of a video summary is to convey the story, and a story could be better understood by analyzing the video from a human perspective. The next section deals with the human aspects that have been considered for the purpose of video skimming.

\subsection{Human Aspect }
As a video summary is a consumable for the human viewer, it is compelling to understand how humans try to get the essence of the video. Modeling semantics, attention, affect and machine learning are a few well known ways of considering the human aspect in computer vision. 

Semantics represent a higher level of understanding, in terms of identifying the entities (objects), context (scenes) and their interactions (events) \cite{chen2009novel}. This higher level of understanding will inherently bring about the possibility of identifying the video segments in line with that of a human viewer which will lead to the generation of human like automatic summaries.

The availability of computational models for visual attention, have paved an appropriate way for incorporating human attention into video summarization algorithms. Attention can be modeled as top down or bottom up \cite{frintrop2010computational}. Video is instrumental in grabbing attention and this continuous attention gives rise to the live experience of the activities of the scene. So modeling of human visual attention, which shows the attentiveness of the viewer seems to be a promising tool for trying out in video summarization.

A goal of skimming could be to have a video skim equivalent to that of the video in terms of affectiveness \cite{7106468} or the kind of emotions/ excitement evoked in the viewer. This aspect is essential for videos containing drama like movies, TV shows and also the news or sports videos, which establishes certain emotions in its viewers. The objective is to make the video skim contain all those affective parts which will give the viewer the thrill of watching the complete video, for example, sports highlights. 

Machine learning is a paradigm that has recently become popular for generating human-like video summaries. They create a model by performing statistical analysis on the given (sufficient) training examples. The created models are efficient in associating the underlying patterns in the data with decisions that can be used to create summaries for new /unseen videos. A learned video skimming model to mimic humans is fundamentally different from the other three because it does not explicitly consider well-known aspects of human's video understanding but statistically estimate them.

The works in semantic category (Section \ref{sssec:semantic}) are grouped into concept detection, entity and interaction and viewer query, as these provide an higher level understanding with respect to human beings. The works in the affect category (Section \ref{sssec:affect}) are grouped into use of facial expression and gaze, and physiological factors, which are the parameters considered for measuring the affectiveness of the video content. The works under attention (Section \ref{sssec:attention}) and hybrid (Section \ref{sssec:hybrid}) modeling are grouped on the basis of video domain. The machine learning (Section \ref{learning}) based works are grouped into classical and deep learning. Finally, we provide a concluding discussion in Section  \ref{humandiscuss}.

\subsubsection{Semantic: }
\label{sssec:semantic}
The semantic concepts deal with high-level human interpretation of the video contents, which cover a wide range  of components such as  objects (e.g., car, airplane), environment (e.g., meeting, desert), events (e.g., people marching),  etc., along with modeling the interactions between them to understand their relationship. Unlike conventional approaches, the emphasis here is on representing the visual concepts through language (like humans) and on deriving the meaning from their interactions. These approaches are usually suited for videos having a rich content structure such as movies and documentaries.  Concept detection based works extract essential concepts from the videos for determining the usefulness of a segment to be part of the summary. Camera movement based semantics can be determined \cite{5582561} for sports summarization. \cite{kannan2015you} identifies a variety of concepts such as a beach, people, flowers, indoors, etc., for ranking the frames and \cite{darabi2015personalized} provides for personalization by adapting summary to viewer preferences of concepts. The works in \cite{tsai2013scene,chen2009novel,lu2013story,7457669} not only identify the concepts but also model the interactions between the prominent entities (any predefined person or object) for determining the representative ones. Viewer query based techniques  \cite{Sharghi2016,sharghi2017query,8099601,xu2008novel,song2015tvsum,7422088} provide for personalization by fine-tuning the summaries to the concepts provided by the user query or some external data like video titles, a text description of video plot, etc.

{\bf Concept Detection. } \cite{vasconcelos1998bayesian} is possibly the first paper to consider semantic content detection for movie summarization based on Bayesian theory. The system is capable of identifying four patterns based semantics in the shots, namely, natural or human-made, action content, facial presence and presence of the crowd. A Bayesian network based inference mechanism is involved in inferring the semantics using motion, skin tone, and texture features. The presence of the semantics in the shots means that it is more important than others and is thus included in the video summary. \cite{cheng2009smartplayer} perform skimming by adjusting the playback speed based on semantic rules, motion analysis and also allows for personalization by learning from user trends. No part of the video is removed, only the playing speed of unimportant skim units is increased. The advantage gained is that the context of the video is not lost and the viewer can understand the complete video. \cite{7547309} provides a framework for surveillance video summarization, where the relationship between objects and events are considered crucial for understanding the context information. The primary contribution of this proposal is to consider local motion regions and their interactions. Correlation graphs are used to learn the spatiotemporal correlations between the features. A sparse coding model \cite{friedman2010note} is used to learn the feature dictionary and any video segment whose features can be represented by the existing features in the dictionary is deemed insignificant. The features of a segment that cannot be represented by the existing dictionary are appended to it. 

In \cite{5582561}, video is segmented based on the rules of TV production \cite{owens2015television}, where the semantics of the game are determined by analyzing the kind of camera view (close-up, medium and far). Close-up are used for game kick-off events while far or medium view is maintained until a goal is scored or defended. The audio features are derived as suggested in \cite{5133759}. Clips within segments are identified using view type and scene type (game or public). Summarization is approached with a resource allocation strategy \cite{6671471}, where the summary time is divided between the clips in the segment. HMM's are used in \cite{6422365} to segment a soccer video into semantically meaningful `play-break' (refer table \ref{videodomains}) sequences. Hue histogram, motion vector length and frame type (I-frame, P-frame and B-frame of MPEG coding) are used for shot boundary detection. Type of camera view (long, medium, close-up, and out-of-field) of the shot is determined by enclosing the players in a bounding box, to measure their size in the video frame. Replay detection is done by identifying the disappearance of TV channel logo. Event type (goal, goal attempt etc.) detection is performed using Bayesian networks. Video summary is formed by applying 0/1 knapsack selection on the play-break sequences with  appropriate weights assigned to each of the sequence.

\cite{kannan2015you} proposes a personalized movie summarization system based on identifying a variety of semantic concepts in the frames using IMARS \cite{natsev2008ibm}, considering similarity among the concepts along with characters appearing in the scenes and shots. \cite{Wang:2014:RSU:2647868.2655013} considers learning the semantic concepts using a classifier and  includes quality (using motion) information for determining important segments. \cite{darabi2015personalized} provides personalization by maintaining user profile that contains concept preferences. \cite{Mei:2013:NSV:2487268.2487269} preserves the semantic meaning/content of the video with reduced storage space and is capable of reconstructing the whole video. Each sub-shot is classified into one of the four classes (object motion, zoom, pan/ tilt  or static) using the affine model \cite{795057}. Each sub-shot is summarized independently using the dominant frame of the video and its metadata. Audio content analysis \cite{Lu2003}, is done to ``partition the audio track into units of 0.5 second and classify each unit among five categories: silence, music, background, pure speech and non-pure-speech'' \cite{Mei:2013:NSV:2487268.2487269}. Parts of the speech signal other than silence are compressed with amr speech codec. The near lossless semantic summary consists of metadata (xml file that captures the motion type, motion duration), compressed key frame and compressed audio.

{\bf Entity and Interaction. } As entities and their interaction (behavior) provide an understanding of the plot of the video, the works \cite{tsai2013scene, chen2009novel,7780487,7457669,lu2013story,Sharghi2016,8099601} identify the critical entities and track them to extract meaningful summaries. The authors of \cite{tsai2013scene} propose movie summarization based on identifying and analyzing the interactions between different roles in the movie by constructing a network of roles existing in each scene. The video stream is segmented into shots using \cite{rasheed2005detection}, followed by facial trajectory extraction and clustering in order to determine the appearance of the same character across the video segments. A character community network is created, which is a graph showing the interactions between various characters of the movie. Video segments which are redundant or that does not contain any interactions between the character are excluded from video skim. Similarly, semantic graph mining is performed in \cite{chen2009novel} for determining important segments in the video.

Authors in \cite{lu2013story} discuss a summarization technique for egocentric videos with the focus on expressing story of the video by modeling the interaction between the entities in different/ consecutive video segments. This work is inspired from the text analysis technique that connects news articles \cite{shahaf2010connecting}. Segmentation is performed by learning the user activities, which are classified using a Support Vector Machine (SVM) classifier. Each of the segments is represented by the visual objects that appear in them. Segment scoring is done by measuring the probability of random walk beginning from one segment and ending on another for each of the objects. The summary is generated by combining the chains of stories based on their scores. Similar idea of video object based summarization using keyframes is proposed in \cite{7780487} for surveillance videos. 

An entity discovery based video skimming approach is proposed in \cite{7457669}. Entity discovery is modeled as tracklet clustering by leveraging temporal coherence using Bayesian non-parametric approach, as the number of entities that will be discovered is unknown. A tracklet is a set of 10-15 frames that contain the same entity. As the identified entities represent semantics, a meaningful summary can be obtained by including the prominent entities in the video. The clusters identified above denote the number of entities in the video, so the process of summarization can be achieved by selecting clusters that cover a large number of entities.

{ \bf Viewer Query. } Here the summarization process is guided by user query, these methods are apt for generating summaries for video search engines.
In~\cite{Sharghi2016}, a probabilistic model, sequential-hierarchical DPP (SH-DPP)  a modified version of Determinantal Point Process (DPP) \cite{kulesza2012determinantal}  is used for this purpose. A video segment that is relevant to the query as well as that is important and diverse throughout the video is selected to be a part of the summary. The user queries were limited to nouns that can be found in the video and are machine detectable. The annotation available within the dataset of \cite{yeung2014videoset} is utilized for identifying the possible nouns for training. Oracle summaries that act as ground truth for a particular query-video (annotations) pair are also used in training SH-DPP. Oracle summaries as suggested in \cite{gong2014diverse} are created by choosing those frames from multiple human summaries which maximally increases the f-measure. A similar approach using DPP is also experimented in \cite{gong2014diverse,sharghi2017query} for performing summarization. Unlike \cite{Sharghi2016}, which used keywords as query, the work in \cite{8099601} provides facility for guiding the summarization process with free-form language description. It extends the sub-modular objectives of \cite{gygli2015video}, by including vision-language objectives. Vision-language embedding is learned using the two branch network of \cite{7780910}, where  visual features are fed to one of the branches and text descriptions to the other. Weights for semantic representativeness and semantic interestingness are also learned along with visual representativeness, visual interestingness and uniformity. This paper also demonstrates the usefulness of low level visual similarity and  semantic similarity in generating good summaries matching the user description.
\cite{xu2008novel} proposes a sports event summarization framework specific to a user query. The user query is first analyzed for keywords and classes, where keywords, identify entities and classes identify the type of query (general/specific, player/team, event/game). Textual description of the corresponding web cast of the game is used to detect events employing a machine learning framework involving HMM. These events are matched to the user query and the relevant portions are considered for summary generation.

\cite{chung2014personalized} uses unsupervised learning of important events in the video based on the user comments while watching the video. All the comments of viewers are synchronized to the player time, the current viewer can view what others have commented on the particular segment of the video. Semantic preference of the viewer is predicted based on semantic user clusters created using comments written by them. Once user cluster is identified, it is further used to generate a personalized summary of episodes which the user is yet to watch. Another unsupervised learning framework that uses video titles to find significant video segments is provided in \cite{song2015tvsum}. It performs co-archetypal analysis  between the video title related images and video segments to find segments representing the visual concepts (semantics). Archetypal analysis \cite{cutler1994archetypal} refers to finding the latent variables that account for the common canonical patterns among the two datasets. Similarly, given a set of scenes of a surveillance video, \cite{7422088} performs Latent Dirichlet Allocation (LDA) \cite{blei2003latent}  to learn the activities in the scenes, LDA generates a set of topics to explain each scene. Scene clustering (spectral clustering) is performed to group scenes with similar activities (topics). A shared representation is obtained for each of the clusters. Summarization is achieved by defining an objective function such that all scenes in each cluster is closer to some scene in the summary (representatives).

{\bf Multiple videos. } Multiple videos are summarized using semantic information in \cite{gao2017event}. First, the shot cliques are identified among multiple videos by grouping near duplicate shots using maximum cohesive subgraph \cite{huang2010mining}. Semantic annotation of each shot clique by a semantic keyword is performed with the help of video title and relevant images mined from the web related to that video title. Shots are temporally ordered by inferring the individual orders from respective videos and estimating the time order for other shots. Shots cliques are selected in such way as to maximize the semantic diversity and coverage, whereas shots from each shot cliques are selected in a way to maximize the smoothness transition between the shots.

\subsubsection{Attention: }
\label{sssec:attention}
Visual attention is the integral part of the `Human Visual System' by which humans tend to understand their surroundings, by attending only to important objects in the scene. Therefore these objects are salient ones and they attract the viewer. Attention while free-viewing a scene usually does not require any semantic understanding and hence attention modeling in general can be separated from semantic modeling. The significance of an image or video frame can be computed based on the salient regions contained in it, so in turn, summarization of the video can be done by selecting the subset of skim units which have a higher number of attention drawing areas. It is assumed that this subset is sufficient to give the viewer the main theme of the video. The latest review of computational visual attention modeling can be found in \cite{borji2013state}.

Movie summarization is explored in \cite{evangelopoulos2008movie, evangelopoulos2009video, evangelopoulos2013multimodal} utilizing video, audio and text saliency scores. The works which are generic with respect to video types, propose ways to model attention phenomena, like the attention model \cite{ma2002model} and perception model \cite{you2007multiple}, which can be applied to any video domain.

{\bf{Movie Videos. } }
\cite{ngo2005video} analyzes the video structure by representing the video as a graph.  The video is partitioned into shots based on the analysis of motion pattern using the 3D spatio-temporal image volumes \cite{Ngo2002}. The obtained shots are represented as nodes in a graph, where edge weights represent similarity between shots, and then normalized cut is used for clustering. A temporal graph of the video clusters is then formed, which is similar to the scene transition graph of \cite{yeung1997video} and is used to model the scenes. Each video cluster being a cluster of shot renders a hierarchical nature of the entire graphical structure. Motion attention values in the scenes and shots within them are then computed using \cite{ma2002model} and selection of segments (Scenes, shots within scenes, and sub-shots within shots) is performed by trimming off unimportant parts at each level.

\cite{evangelopoulos2008movie, evangelopoulos2009video, evangelopoulos2013multimodal} suggest a movie summarization method using audio, video and text saliency. Audio saliency determined by linear combination of maximum  average teager energy, mean instant amplitude and mean instant frequency. Video saliency is determined by extracting color, intensity and orientation for each frames followed by decomposing the video into different spatio-temporal scales. The saliency is obtained by measuring the rarity of the features in time along with the consistency of the rarity across scales. Text analysis is performed on the movie sub-titles.  Each of the words is given importance scores based on Parts of Speech (POS)~\cite{schmid2013probabilistic} tagging.  The three saliency scores are fused and normalized for determining segment importance. \cite{7351630} is extension of previous technique with machine learning based shots selection. K-nearest neighbor classifier is used to determine whether a frame is salient or not.

{\bf{Generic Methods. }}
In \cite{ma2002model}, a motion attention model is proposed for video summarization. Here motion saliency map is found out by fusing the intensity, spatial coherence and temporal coherence maps obtained from the motion vector field encoded in MPEG. The saliency map of every frame is then processed to highlight the attention grabbing regions. The frames are given importance proportional to the attention area in it. Then motion attention indicator curve over time is formed and the video skim is generated by selecting segments above certain threshold. This work is further extended in  \cite{ma2002user}, \cite{ma2005generic} to create a framework for user attention by including curves for static attention  \cite{itti1998model}, face attention, camera motion and audio attention.

In \cite{you2007multiple}, a perception model is created, which tries to mimic human understanding of videos based on multiple visual cues, for video summarization.  The modeling is based on motion, contrast, special scenes and video rhythm. Local motion, the one caused by object motion in the video, attracts human attention. The luminance difference between a local motion region and others constitute the contrast. For a frame with negligible motion the contrast is determined by considering the difference between two dominant luminance regions.  Special scenes are those consisting of human face and captions. Rhythm is calculated based on the average energy of macro blocks in a video segment. Importance is associated at the frame, shot and scene level depending on the above four attributes used in the perception modeling. The paper suggests a linear fusion of the importance values for videos with simple content and an appropriate non-linear fusion for videos where frequent shot switching takes place.  It provides for a scalable length video skims by using perceptual importance curve. The curve can be cut at different thresholds to generate different length skims.

\subsubsection{Affect: }
\label{sssec:affect}
These approaches consider measurement of the emotional impact of the video. In most methods, the affectiveness is quantified through user behavior (non intrusive  technique) while watching the video based on measuring some physiological  parameters. These methods involve passive participation of users in generating the summaries. These are designed to overcome the lacuna's of fully automatic systems available for generating skims, which incorporate semantics and have dependencies on the video domain. 

The works are grouped on the basis of the parameters considered for measuring affectiveness. While facial expression and gaze are used in \cite{peng2009user,joho2009exploiting,katti2011affective,xu2015gaze} to identify viewer emotions, physiological parameters are measured in ~\cite{money2010elvis,peng2011editing} for the same. Emotions are determined by some other intuitive ways as well, such as user tweets in \cite{Tang:2012:ECS:2207676.2208622} and from expressions of the lead characters of the video in \cite{ xiang2011affect}.

{\bf Facial expression and gaze. } Among the methods, use of facial expression and gaze has been considered widely. \cite{peng2009user} is one such method, which tackles the issue of home video (user video) skimming. A human experience model is built by combining facial expression and eye movements, as humans understand a scene by looking at particular parts of the video. The eye gaze and facial expression of the viewer can be used in deciding whether they are able to grab the meaning. In the paper, fixation for a long time is considered to represent high viewer interest, whereas more saccade implies lesser viewer interest. Whereas facial expression is identified for finding the viewers like/dislike about the video clip. Facial emotion is considered, in which positive and negative  expressions are used to understand the viewers' intention. A weighted fusion of importance scores due to motion, facial expression and eye movement is used for assigning a score to a frame/shot.  The weights are determined by the kind of shots under analysis, i.e., in motion shots, camera and object motion is more indicative of importance than eye movement. In static shots, it is the opposite. Selection of segments is based on the score until the summary length constraint is satisfied.

In \cite{joho2009exploiting}, facial expressions are used for extracting the affective segments of the video. Motion units are used to estimate the 3D face motion from 2D points extracted from faces. Prominence of expression is considered and classified into three categories, namely, nil, low and high. Frequency of change in facial expression is also considered along with prominence to select important video segments (shots). 

\cite{katti2011affective} uses pupillary dilation (PD) in eyes while viewing videos for generating their summaries, as it is a means of identifying the interest and engagement of the viewer. Standard content analysis is performed for video segmentation. Important segments are then identified by measuring the deviation on the larger side from the mean PD values, based on which the skim is formed. To generate storyboards, eye gaze and PD is considered for identifying keyframes of interest that are compiled together. \cite{yoshitaka2012personalized} considers eye movement and remote control usage to determine the preferences of the viewer. Important segments from the perspective of the viewer is identified when the viewer forwards or replays a particular segment of the video. In addition, user's gaze fixation gives information on the users attention. Based on these two factors, selection of important scenes for summarizing the video is formed. The summarization is demonstrated for sports videos. \cite{xu2015gaze} uses human gaze data to generate a personalized summary for egocentric video. Segmentation of video is performed using the gaze information. Maximization of relevance (segment importance) and diversity (segment uniqueness) of the segments in the summary to be formed is considered along with importance based on gaze.

{\bf Physiological Factors. } \cite{money2010elvis} measures physiological responses for identifying important segments in the video. Electro-Dermal Response (EDR), Respiration Amplitude (RA), Heart Rate (HR), Blood Volume Pulse (BVP), Respiration Rate (RR) and Respiration Volume (RV) are used. EDR measures the electrical conductivity of the skin, through sweat produced by glands. Higher the EDR higher the arousal. RA is used to indicate arousal  and valence levels. RR is increased number of breaths per minute implying increased arousal. BVP measures the volume of blood  pumped into the body, this is used to measure arousal. HR is used as an indicator of valence. First, the physiological responses are captured and standardized. Standardization is  required as different sensors have different sampling rates and quantization schemes. Second, the values are averaged using sliding window, where the window size is determined by using a minimum video segment length provided by the viewer. All the above importance measure is combined into a single value called  entertainment  value. Segments are identified based on the entertainment value to be included in the summary. 

The authors of \cite{peng2010real} and \cite{peng2011editing} suggest  a psychometric approach used for summarizing video in an easy and efficient manner. Here the summarization is based on humans' actions and analyzing the psychological states rather than analyzing visual/aural  variations. The authors propose interest meter (IM), a score derived from the combination of attention and emotion model. Attention is measured  by blink detection, saccade detection and head motion detection. Emotion states are measured using facial expression recognition. A probabilistic SVM classifier is trained to learn happy and neutral emotions. A fuzzy logic based information fusion is done to combine the two importance scores based on attention and emotion. 

{\bf Other Approaches. }
In \cite{Tang:2012:ECS:2207676.2208622}, TwitInfo algorithm \cite{Marcus:2011:TAV:1978942.1978975} is used to collect real time feedback of viewers from Twitter. A graph of number of tweets over time is plotted. The period of time receiving more tweets correspond to interesting events, these are used to form the summary. Although such summary does not guarantee coverage of all the interesting parts of the program but it reflects the viewers emotions.

\cite{xiang2011affect} performs a sparsity based affect analysis, where the affective content of a shot is identified by sparse learning. Human face  recognition is also used for identifying significant/lead persons in the video. These affect and face labels are assigned to each shot. Using a modified term frequency (TF-IDF) technique on affect labels, the emotional tone of the video is obtained, which represents the emotions mostly felt by audience. Local main character is defined as the main character in a video and global main characters are defined as the main characters from a collection of videos on the same theme. Importance weights for shots are calculated  based on the emotional tone, and the local and global main characters. These weights are used to perform summarization. In \cite{xiang2011affect}, their affect based personalized video presentations as shorter versions of long videos have been found effective for consumption. This is so, as such personalization can be appropriately customized for social groups such as family and acquaintances.

\subsubsection{Hybrid: }
\label{sssec:hybrid}
In this section, we discuss some techniques which uses a combination of the above `Human Aspects' (semantic, attention and affect) to derive summaries. The works on movie summarization \cite{ren2010semantic,dong2014iteratively,7362797} perform subtitle based semantic topic identification along with salience and affective analysis for choosing video segments. Egocentric tour videos are summarized in \cite{Varini:2015:EVS:2733373.2806367} by inferring tourist behavior from GPS information. Under generic methods, \cite{mehmood2016divide} investigates fusing EEG signal with attention scores.

{\bf{Movie Videos. }}
The authors of \cite{ren2010semantic} suggest movie summarization based on internal and external textual descriptions. Internal descriptions are available with the subtitles / closed captions and the external description is given by users. Plot of movies is taken from Wikipedia and compared with both descriptions to estimate the content topic distribution using LDA. The Wikipedia plot contains all semantic details and is helpful in bridging the gap between generic user statements and noisy subtitles. Video segments are identified based on prominent topics.  Segments in the content topics are scored using audio and video saliency based on human attention modeling, followed by selecting top 20\% important segments from each topic for skim generation. \cite{dong2010video} and \cite{dong2014iteratively} use semantic correlation between video contents and corresponding text to determine a segment's importance. Each video segment is identified by a key frame. Semantic features that represent high level concepts are extracted from the key frame of each segment using Bag-of-Features approach \cite {jiang2007towards}. The similarity of the extracted concept to the major words obtained from subtitles (around the key frame) is measured using WordNet::Similarity \cite{pedersen2004wordnet}, with higher similarity indicating higher semantic coherence. Apart from semantic features, motion and face attention features are also considered. The attention features are obtained using the user attention model of \cite{ma2002model}. A linear fusion of importance based on semantic coherence, and face and motion attention presence is performed to obtain an importance curve. For the importance based on semantic coherence, concept filtering is considered by thresholding on the coherence assuming that not all semantic features would be useful for summary generation. Video skimming is shown considering news, documentaries and movies. 

Movie summaries are created in~\cite{7362797} by a combination of audio saliency and affective ratings of words in the subtitles. As particular words impact the viewer of the movie, such a technique works well to access the importance of the video segment.  In \cite{xu2016fast}, video segments are scored on the basis of emotion, semantic and quality (using motion stability and lighting) to determine important segments. \cite{yuan2011video} proposes to identify salient concepts in the video segments.

{\bf{Egocentric Videos. }}
\cite{Varini:2015:EVS:2733373.2806367} proposes to provide a custom summary for a cultural tour video by inferring important segments by analyzing tourists' visiting behavior. The preferences of the tourists are hidden in the way s/he does their tour. The motion of the camera wearer is classified into `body still' and `body moving' classes based on head movement information. The GPS position of the camera wearer provides information on the stops made at a particular location during the visit. Utilizing the tourist stop location, relevant semantic topics are extracted from the segments. The topics are expanded using DBpedia~\cite{Auer:2007:DNW:1785162.1785216} (dataset containing topic information) to connect similar topics and then images are selected for these topics. These are used for training semantic classifiers using BOW. Shots to be included in the summary are chosen based on the classification scores and visual diversity. The work is extended in \cite{7931584}, where video segmentation is done by learning homogeneous behavior pattern. The pattern is learned using a CNN into the following classes: attention, looking around, walking, running, on wheels and wandering. Shot narrativity score is determined using personalized page rank \cite{kamvar2003extrapolation}. Segments are selected based on maximizing an objective function considering behavior, semantic concepts and narrativity score.

{\bf{Generic Methods. }}
In \cite{han2011personalized}, the authors suggest that the semantic information is better understood by viewers. So, by using a set of key frames provided by a viewer, the system generates the summary for that viewer based on visual similarity to the given set of frames. 
The availability of affordable bio-sensors have motivated \cite{mehmood2016divide} to use EEG brain signals to identify the affective content of the video. For example, an Emotiv EPOC headset \cite{epoc} can measure a viewer's neuronal response in real time. Inter-frame chi-square histogram difference is used for shot boundary detection. A fused importance curve is formed using EEG together with audio and video attention for the skimming process. It is suggested that this method can be extended to perform skimming based on affect detection in shots.

\subsubsection{Machine Learning: }
\label{learning}
These techniques utilize example summaries created by humans to learn a model that is capable of choosing the video frames into the summary. The two sub-groups are classical learning  and deep learning. Classical techniques use hand-crafted features for learning the model whereas deep learning involves Convolutional Neural Networks (CNN) that extract deep features \cite{donahue2014decaf} and also automatically perform feature selection as the learning progresses. Both of them rely on accurately labeled ground truth data, but deep learning techniques require a significant larger amount of training data as their capability of developing an end to end (from input to desired output) learned application requires substantially more parameters to be tuned.   

Classical learning and deep learning techniques are mostly supervised /weakly supervised by using ground truth /additional cues to learn the model. Learning from relevant photos \cite{khosla2013large}, or from videos belonging to a specific event category like changing the tire, parkour, etc., can be done as they possess a common structure among them \cite{potapov2014category,zhang2016summary,panda2017weakly}. Utilizing raw videos along with edited videos for relative ranking can be done to avoid the task of ground truth creation \cite{sun2014ranking,gygli2016video2gif}. Supervised optimization methods can be employed for creating generic models which can serve any video domain \cite{gygli2015video,7904630}. In addition, there are specific deep learning models that aim at exploiting temporal dependencies within the video using Long Short Term Memory (LSTM) network \cite{zhang2016video,zhao2017hierarchical} and perform unsupervised learning using autoencoder network \cite{yang2015unsupervised,8099801} that learn the relative importance among the video segments through the encoding error accumulated in the network.

{\bf{Classical Learning. }} The works under this category can be further sub-grouped into unsupervised, weakly supervised and supervised machine learning approaches. However, classical weakly supervised and unsupervised learning have seldom been used in video skimming \cite{khosla2013large}. \cite{khosla2013large} uses web images in guiding the summarization process for user videos (their dataset consists of cars and trucks videos).  Canonical viewpoints are identified from web images belonging to the same object class as that of the video. A viewpoint refers to the camera angle for capturing the object in a maximally informative way. The viewpoints are learned using an SVM classifier. Identifying and selecting such viewpoints from the video will be sufficient to summarize the video. Although canonical viewpoints of objects can be found by crawling images from the Internet, this may not be possible for all scenarios. For example, in cooking videos, the images of different dishes while cooking might look the same from different viewpoints.

A category specific video summarization approach is suggested in \cite{potapov2014category}. Video segmentation is performed using change point detection \cite{kay1998fundamentals} between the frames, which not only identifies the shot boundaries but also the sub-shots. While a binary SVM classifier is trained for each video category to perform video classification, a linear SVM trained on human produced data is used to score the video segments relative to the video category. High scoring segments from the same category videos are selected to form the skim constrained to the required summary length.
\cite{zhang2016summary} suggests a non-parametric machine learning  approach for video summarization, where the category-wise summary structure is learned from training videos with available human created summaries. The idea is based on structural similarity among the videos belonging to the same category. The authors pointed out that use of similarity between the model learned from human created summaries and a test video will not work, as it may require large amounts of data to learn for all possible cases. Instead, the authors do video structure learning from a selected subset of shots and compare this learned structure with a selected subset of shots in a test video using sub-shot or frame level similarity. The DPP of \cite{kulesza2011learning} is then used to create the video summary based on the similarities.

 \cite{lin2015summarizing} proposes an online highlight detection framework on the basis of context information. A structured SVM is used to learn the context and highlight from training videos such that relevant segments are getting higher scores than others. Loss function/objective function is designed to jointly consider context and highlight, such that the most probable context is determined using the available video frames followed by applying the context specific model for highlight selection.

\cite{gygli2015video} suggests the use of a supervised model for learning and jointly optimizing different objectives, such as uniformity, representativeness and interestingness to generate video  summaries. Sub-modular functions are defined for each of the objectives. Interestingness of a segment is found by taking the sum of interestingness of each of the individual frames in the segment. Representativeness measures how well the selected segments convey the contents of the original video. This is modeled as selecting k most interesting segments from the video such that the sum of squared error between the segment medoids and data points is minimized. Finally uniformity factor preserves the temporal coherence of the original video, preventing frequent jumps or many adjacent segments in the skim, which may lead to redundancy. 

\cite{sun2014ranking} generates highlights by performing pairwise learning between the segments of raw and edited videos, such that highlight segments are always scored higher than non-highlight segments. The video is segmented into fixed size (100 frames) segments. A latent linear ranking SVM is trained to rank the segment depending on their suitability to become a highlight. Training data is harvested through mining YouTube, i.e., raw video and edited video belonging to the same category are used for learning. A generalized framework is proposed in \cite{7904630} to summarize either edited or raw videos. The videos are scored on criteria such as: importance, representative, diversity, and storyness. Since the raw videos and edited videos consider the above properties in different proportions, generalization is achieved by defining the weights for each of these videos independently by providing mixing coefficients. The training set consists of both edited and raw videos. Each video in the training set is provided with two coefficients, one for the edited and one for the raw. The weights are decided based on the given video's similarity to that of training set. The average of the mixing coefficients determine the individual weights given to each of the above properties for summarizing the given video. 

{\bf{Deep Learning. }} The works here are further sub-grouped into unsupervised, weakly supervised and supervised.

\underline{\textit{Unsupervised}: } \cite{yang2015unsupervised} proposes an unsupervised approach for generating highlights by learning from only edited videos. An Auto-Encoder Recurrent Neural Network (AE-RNN) is trained on a set of videos belonging to a particular category like swimming, skating, etc. Edited videos belonging to such classes are easily available on social media websites, which essentially capture the highlight. AE-RNN is combined with bi-directional LSTM for tracking temporal variations in the video. The AE-RNN learning is augmented with a shrinking loss function in order to reduce the influence of  noise that is naturally present in web videos. Given a testing video, segmentation is performed using kernel temporal segmentation (KTS) \cite{potapov2014category} with an additional constraint to restrict the segment length to [48,96] frames. The model outputs the confidence scores (encoding error) for the segments, and those which have least encoding error are selected to be the highlight of the given video. 
\cite{8099801} performs unsupervised learning using adversarial LSTM network. The idea is to train a variational auto-encoder for unsupervised learning of video features to be extracted by selecting a suitable set of frames from the video. Since the method is unsupervised, a discriminator is used as in a Generative Adversarial Network (GAN) setting. The LSTM decoder reconstructs the video from the selected frames, which is fed to the discriminator. The discriminator attempts at classifying the input into original or summary classes.

\underline{\textit{Weakly supervised}: } \cite{panda2017weakly} suggests an intermediate way between supervised and unsupervised methods i.e., weakly supervised method. Given a set of videos belonging to a category, 3D CNN features are used to learn the video category.  Depending on the predicted video category, segments from the video are chosen during back propagation. Segment scoring is done on the basis of gradient information obtained from the network. The authors suggest different ways for training considering the limited size of available datasets namely, cross dataset training, web crawled videos and augmentation methods for providing sufficient training data. \cite{ho2018summarizing} argues that techniques applied on egocentric video summarization by training on user videos is not effective and suggests a semi-supervised approach by utilizing annotated egocentric video, annotated user video along with unlabeled egocentric videos and performs learning on a common feature space.

\underline{\textit{Supervised}: }
In \cite{yaohighlight} deep learning is used to discern between the highlight and non-highlight video segments of egocentric video. The approach performs pairwise relationship learning between the video segments in order to incorporate relative preferences among them. The video stream is processed spatially and temporally using two deep convolution neural networks to assign the relative preference score to the segments. The summary is generated by appending the segments with highest scores. Similarly, \cite{gygli2016video2gif} creates GIFs from videos. ``A GIF is short, entirely visual with no sound, expresses various forms of emotions, and sometimes contains unique spatio-temporal visual patterns that make it appear to loop forever'' \cite{gygli2016video2gif}. They capture the single most interesting moment in the video. The videos are segmented using the approach of \cite{song2015tvsum}. Ground Truth GIF's are aligned with their corresponding videos to find the part of the video included in the GIF. Positive segments are those that are part of GIF, while negative are not. Each segment is ranked based on their suitability for forming GIF. This ranking is determined by a novel huber loss function \cite{huber1964robust} to deal with noisy data that also incorporates the quality/popularity of the GIF. The ranking score is such that a positive segment gets a better score than a negative segment. Even though the technique is proposed for GIF, it generalizes well for highlight detection. \cite{del2018phd} provide for personalization of GIF's by using two models, one to learn from the aggregated history of all users and the second for personalization.

In \cite{zhang2016video}, the authors suggest LSTM for learning the temporal difference between similar segments. LSTM is combined with DPP to improve the subset selection among similar segments. To overcome the issue of unavailability of large annotated datasets for training, domain adaptation \cite{sun2016return} is done to include heterogeneous videos from different datasets into a common format. The proposal is shown to work well on videos having a gradual change in its content. In order to handle long-range temporal dependency (more than 80 frames) \cite{zhao2017hierarchical} proposes Hierarchical-RNN. The first layer is an LSTM whereas the second layer is a bi-directional LSTM, short-range dependencies are exploited by first and long-range by the second. The confidence scores given by the second is used to decide on the segment importance. \cite{wei2018video} uses an LSTM network to learn the mapping of a frame's visual content to its corresponding text descriptions. The video is segmented using KTS \cite{potapov2014category} to derive shots, which are combined to form 3/5 segments per video. This is done based on an observation that the evolution of events in each of the user videos can be divided into beginning, climax and ending phases. Each segment is annotated by three users to describe the event accurately, i.e., by using generic words to specific words. The LSTM learns by minimizing the distance between the ground truth description and the predicted description. \cite{ji2017video} performs summarization using encoder-decoder (LSTM network) setting with additional attention (attention model of \cite{luong2015effective}) based weights provided in the decoder. This provision enables giving scores to different video frames such that not only redundancy is removed but also content saliency is preserved.

\subsubsection{Discussions: }
\label{humandiscuss}
Semantic techniques try to understand the viewer preferences in terms of semantic concepts as well as by segmenting the video into semantically meaningful parts for further processing. Attention analysis strives to assign saliency scores for each of the video frames by processing various modalities thoroughly to evaluate the different components in a video frame. However, as pixel-wise saliency computation is performed in every frame, it is computationally intensive. The affect based techniques presents a personalized way of generating video summaries by taking viewer response. Although they have been found to be very effective in capturing the human propensities, most of them require the viewer to watch the entire video before the summarization is performed. These techniques require active user involvement, and as viewer physiological states are measured, it might be difficult to ensure the viewer is completely focused on the video. Therefore, one might prefer to infer viewer emotion states from the video itself without viewer's involvement. Deep learning techniques are being employed for video summarization that consider the temporal dependencies among the video frames for determining importance. Hybrid approaches are those which use multiple human aspects for summary generation. They intend to verify active viewer involvement by considering other clues from the video (semantic concepts in subtitles and their relevance to the video concepts). Understanding which of the human aspects has a larger impact on the summarization is a problem that should be extensively studied. It is imperative that human beings perform low, mid and high-level processing \cite{frintrop2010computational} for understanding the scenes in the video, therefore all the above aspects do play a role in concluding whether a video segment is interesting or not. Based on studies on the impact of these aspects, an appropriate fusion mechanism needs to be considered.

We give an overview of most of the approaches discussed in this survey in Table \ref{tableAV}. Keeping in line with our taxonomy and generic framework, we outline the kind of segmentation technique used, the kind of skim unit selected, modalities utilized and the video skimming technique employed.

\begin{center}
\tiny
\begin{longtable}{|l|l|cccc|c|cccc|c|cccccccc|}
\caption{Details of different video skimming contributions in terms of video skim unit, modalities, datasets and skimming technique. The abbreviated category are: Af=Affect, At=Attention, S=Semantic, ML=Machine Learning, Mo=Mono-view, Mv=Multi-view, O=Online. Abbreviated segmentation techniques are: CD=color histogram difference, ID=intensity difference, G=graph based, MT=motion based, CA=content analysis, E=external data, T=tool, A=audio, L=learning, VP=video production rules, F=fixed size. *Segment: refer to any set of consecutive frames.}
\label{tableAV} \tabularnewline
\hline
 &  & \multicolumn{4}{c|}{\textbf{Video Skim Unit}} & & \multicolumn{4}{c|}{\textbf{Modality Usage}} & \textbf{} & \multicolumn{8}{c|}{\textbf{Video Skimming Technique}} \tabularnewline \hline
\multicolumn{1}{|c|}{\textbf{\rotatebox[origin=c]{90}{Reference}}} & \multicolumn{1}{c|}{\textbf{\rotatebox[origin=c]{90}{\begin{tabular}[c]{@{}l@{}}Category \\(see Fig. \ref{fig:reviewclassification})\end{tabular}}}} & \multicolumn{1}{c}{\textbf{\rotatebox[origin=c]{90}{Frame}}} & \multicolumn{1}{c}{\textbf{\rotatebox[origin=c]{90}{Shot}}} & \multicolumn{1}{c}{\textbf{\rotatebox[origin=c]{90}{Scene}}} & \multicolumn{1}{c|}{\textbf{\rotatebox[origin=c]{90}{Segment*}}}&\multicolumn{1}{c|}{\textbf{\rotatebox[origin=c]{90}{\begin{tabular}[c]{@{}l@{}}Segmentation \\Technique\end{tabular}}}} & \multicolumn{1}{c}{\textbf{\rotatebox[origin=c]{90}{Video}}} & \multicolumn{1}{c}{\textbf{\rotatebox[origin=c]{90}{Audio}}} & \multicolumn{1}{c}{\textbf{\rotatebox[origin=c]{90}{Text}}} & \multicolumn{1}{c|}{\textbf{\rotatebox[origin=c]{90}{ \begin{tabular}[c]{@{}l@{}}External \\Data\end{tabular}}}} & \multicolumn{1}{c|}{\textbf{Datasets Used}}& \multicolumn{1}{c}{\textbf{\rotatebox[origin=c]{90}{\begin{tabular}[c]{@{}l@{}}Fusion/\\Selection\end{tabular}}}} & \multicolumn{1}{c}{\textbf{\rotatebox[origin=c]{90}{Clustering}}} & \multicolumn{1}{c}{\textbf{\rotatebox[origin=c]{90}{Graph}}} & \multicolumn{1}{c}{\textbf{\rotatebox[origin=c]{90}{Optimization}}} & \multicolumn{1}{c}{\textbf{\rotatebox[origin=c]{90}{Dictionary}}} & \multicolumn{1}{c}{\textbf{\rotatebox[origin=c]{90}{\begin{tabular}[c]{@{}l@{}}Classical\\Learning\end{tabular}}}} & \multicolumn{1}{c}{\textbf{\rotatebox[origin=c]{90}{Fast-Forward}}}& \multicolumn{1}{c|}{\textbf{\rotatebox[origin=c]{90}{Deep Learning}}} \\ \hline

\endfirsthead
\multicolumn{20}{c}{{\bfseries \tablename \thetable{}-- continued from previous page}}\\
\hline
\textbf{} & \textbf{} & \multicolumn{4}{c|}{\textbf{Video Skim Unit}} & & \multicolumn{4}{c|}{\textbf{Modality Usage}} & \textbf{} & \multicolumn{8}{c|}{\textbf{Video Skimming Technique}} \\ \hline
\multicolumn{1}{|c|}{\textbf{\rotatebox[origin=c]{90}{Reference}}} & \multicolumn{1}{c|}{\textbf{\rotatebox[origin=c]{90}{\begin{tabular}[c]{@{}l@{}}Category \\(see Fig. \ref{fig:reviewclassification})\end{tabular}}}} & \multicolumn{1}{c}{\textbf{\rotatebox[origin=c]{90}{Frame}}} & \multicolumn{1}{c}{\textbf{\rotatebox[origin=c]{90}{Shot}}} & \multicolumn{1}{c}{\textbf{\rotatebox[origin=c]{90}{Scene}}} & \multicolumn{1}{c|}{\textbf{\rotatebox[origin=c]{90}{Segment*}}}&\multicolumn{1}{c|}{\textbf{\rotatebox[origin=c]{90}{\begin{tabular}[c]{@{}l@{}}Segmentation \\Technique\end{tabular}}}} & \multicolumn{1}{c}{\textbf{\rotatebox[origin=c]{90}{Video}}} & \multicolumn{1}{c}{\textbf{\rotatebox[origin=c]{90}{Audio}}} & \multicolumn{1}{c}{\textbf{\rotatebox[origin=c]{90}{Text}}} & \multicolumn{1}{c|}{\textbf{\rotatebox[origin=c]{90}{ \begin{tabular}[c]{@{}l@{}}External \\Data\end{tabular}}}} & \multicolumn{1}{c|}{\textbf{Datasets Used}}& \multicolumn{1}{c}{\textbf{\rotatebox[origin=c]{90}{\begin{tabular}[c]{@{}l@{}}Fusion/\\Selection\end{tabular}}}} & \multicolumn{1}{c}{\textbf{\rotatebox[origin=c]{90}{Clustering}}} & \multicolumn{1}{c}{\textbf{\rotatebox[origin=c]{90}{Graph}}} & \multicolumn{1}{c}{\textbf{\rotatebox[origin=c]{90}{Optimization}}} & \multicolumn{1}{c}{\textbf{\rotatebox[origin=c]{90}{Dictionary}}} & \multicolumn{1}{c}{\textbf{\rotatebox[origin=c]{90}{\begin{tabular}[c]{@{}l@{}}Classical\\Learning\end{tabular}}}} & \multicolumn{1}{c}{\textbf{\rotatebox[origin=c]{90}{Fast-Forward}}}&\multicolumn{1}{c|}{\textbf{\rotatebox[origin=c]{90}{Deep Learning}}} \\
\hline

\endhead

\hline \multicolumn{20}{|r|}{{Continued on next page}}\\ \hline
\endfoot

\hline 
\endlastfoot

\cite{taskiran2006automated} & Mo &  & \checkmark &  &  & T &  & \checkmark &  &  &  & \checkmark &  &  &  &  &  &  &\\
\cite{peng2008aesthetics} & Mo &  & \checkmark &  &  & CD & \checkmark & \checkmark &  &  &  & \checkmark &  &  &  &  & & &  \\
\cite{laganiere2008video} & Mo & \checkmark &  &  &  &  & \checkmark &  &  &  &  & \checkmark &  &  &  &  &  & & \\
\cite{4907069} & Mo &  & \checkmark &  &  & CD & \checkmark &  &  &  &  &  & \checkmark &  &  &  &  &  &\\
\cite{5133759} & Mo &  &  &  &  & A &  & \checkmark &  &  &  & \checkmark &  &  &  &  &  &  &\\
\cite{gao2009dynamic} & Mo &  & \checkmark & \checkmark &  & CD+G & \checkmark & \checkmark &  &  &  &  & \checkmark & \checkmark &  &  &  &  &\\
\cite{5993544} & Mo &  & \checkmark &  &  & L & \checkmark & \checkmark &  &  &  & \checkmark &  &  &  &  & \checkmark &  &\\
\cite{li2011static} & Mo &  &  &  & \checkmark & A & \checkmark & \checkmark & \checkmark &  &  & \checkmark &  &  &  &  &  &  &\\
\cite{cong2012towards} & Mo & \checkmark &  &  &  &  & \checkmark &  &  &  & \cite{loui2007kodak} &  &  &  &  & \checkmark &  & & \\
\cite{6671471} & Mo &  & \checkmark &  &  & VP & \checkmark &  &  &  & \cite{blunsden2010behave,jiastdataset} &  &  &  & \checkmark &  &  & \checkmark &\\
\cite{dang2014heterogeneity} & Mo & \checkmark &  &  &  &  & \checkmark &  &  &  & \cite{loui2007kodak} & \checkmark &  &  &  &  &  &  &\\
\cite{gygli2014creating} & Mo &  &  &  & \checkmark & MT & \checkmark &  &  &  & \cite{gygli2014creating} & \checkmark &  &  & \checkmark &  & \checkmark &  &\\
\cite{7368910} & Mo &  &  &  &  &  &  &  &  & \checkmark & \cite{7368910} & \checkmark &  &  &  &  &  & & \\
\cite{del2017active}	&Mo&	& & &\checkmark	&ID+MT&	\checkmark& & & &\cite{lee2012discovering,del2017active}& & & &	\checkmark& & &&\\
\cite{hong2009event} & Mv &  & \checkmark &  &  & MT+L & \checkmark &  &  &  &  & \checkmark &  &  & \checkmark &  &  &  &\\
\cite{fu2010multi} & Mv &  &  &  & \checkmark & MT & \checkmark &&&& \cite{fu2010multi} & &\checkmark & \checkmark & \checkmark & & &&\\
\cite{hong2011beyond} & Mv &  & \checkmark &  &  & MT+L & \checkmark &  &  &  &  & \checkmark &  &  & \checkmark &  &  &  &\\
\cite{Zhang:2012:MSS:2155555.2155565} & Mv &  & \checkmark &  &  & E & \checkmark &  &  & \checkmark &  & \checkmark &  &  &  && && \\
\cite{wang2012event} & Mv & \checkmark &  &  &  &  & \checkmark& & & \checkmark &  & \checkmark & \checkmark &  & \checkmark &  &  &&  \\
\cite{7121011} & Mv &  & \checkmark &  & & MT & \checkmark & & & & \cite{fu2010multi,6838965} &  & \checkmark & \checkmark &  & &  &&  \\
\cite{6838965} & Mv & \checkmark &  &  &  &  & \checkmark &  &  &  & \cite{fu2010multi,6838965} & \checkmark & \checkmark &  &  &  & &&  \\
\cite{chu2015video} & Mv &  &  &  & \checkmark & ID & \checkmark &  &  &  & \cite{chu2015video} &  &  & \checkmark &  &  &  & & \\
\cite{7532345} & Mv & \checkmark &  &  &  &  & \checkmark &  &  &  & \cite{fu2010multi} &  &  &  & \checkmark & \checkmark &  &&  \\ 
\cite{7952384}&	Mv	& & & &				\checkmark&	ID&	\checkmark		& & & &  & & & &\checkmark	& & &&\\
\cite{6909934}&  Mv &		\checkmark& & & & &\checkmark&& & & & & &	\checkmark& & & &&\\
\cite{8099938}&Mv& & & &\checkmark&ID& \checkmark& & & &	\cite{chu2015video,song2015tvsum}& & & &\checkmark&& &&\\
\cite{ji2019query}&{\color{blue}Mv}&&\checkmark&&&G&\checkmark&&&\checkmark&\cite{song2015tvsum}&&&&&\checkmark&&&\\	
\cite{Valdes:2008:BTB:1463563.1463588} & O &  &  &  & \checkmark & F & \checkmark &  &  &  &  &  &  & \checkmark &  &  &  & & \\
\cite{almeida2013online} & O & \checkmark &  &  &  &  & \checkmark &  &  &  &  &  & \checkmark &  &  &  &  & & \\
\cite{zhao2014quasi} & O &  &  &  & \checkmark & F & \checkmark &  &  &  &  &  &  &  &  & \checkmark &  &  &\\
\cite{7532342} & O &  & \checkmark &  &  & F & \checkmark &  &  &  & \cite{gygli2014creating} & \checkmark &  &  &  & \checkmark &  & & \\ 
\cite{8099680}&O& & & &\checkmark&	MT&\checkmark& & & &\cite{gygli2014creating, song2015tvsum}& & & &\checkmark	&&&&\\	
\cite{chen2009novel} & S &  & \checkmark &  &  & CD & \checkmark & \checkmark &  &  &  &  &  & \checkmark &  &  &  &  &\\
\cite{cheng2009smartplayer} & S &  &  &  & \checkmark & CD+MT & \checkmark &  &  &  &  &  &  &  &  &  &  & \checkmark& \\
\cite{5582561} & S &  & \checkmark &  &  & VP & \checkmark & \checkmark &  &  &  &  &  &  &  &  &  & \checkmark &\\
\cite{dong2010video} & S & \checkmark &  &  &  &  & \checkmark &  & \checkmark &  &  & \checkmark &  &  &  &  &  &  &\\
\cite{tsai2013scene} & S &  & \checkmark & \checkmark &  & G & \checkmark &  &  &  &  &  & \checkmark & \checkmark &  &  &  & & \\
\cite{Wang:2014:RSU:2647868.2655013} & S &  &  &  & \checkmark & F & \checkmark &  &  &  & \cite{Jiang:2011:CVU:1991996.1992025} & \checkmark &  &  &  &  & \checkmark &  &\\
\cite{6422365} & S &  & \checkmark &  &  & MT+CH & \checkmark &  &  &  & \cite{6422365} &  &  &  & \checkmark &  & \checkmark & & \\
\cite{chung2014personalized} & S &  &  &  &  &  &  &  &  & \checkmark &  &  & \checkmark &  &  &  & \checkmark &  &\\
\cite{song2015tvsum} & S &  & \checkmark &  &  & MT & \checkmark &  &  & \checkmark & \cite{gygli2014creating,song2015tvsum} &  & \checkmark &  &  &  &  &  &\\
\cite{darabi2015personalized} & S &  &  & \checkmark &  & T & \checkmark &  &  &  &  &  &  &  &  &  & \checkmark &  &\\
\cite{kannan2015you} & S &  & \checkmark & \checkmark &  &T& \checkmark &  & \checkmark &  &  & \checkmark &  &  & \checkmark &  &  & &\\
\cite{lu2013story} & S &  &  &  & \checkmark & L & \checkmark &  &  &  & \cite{lee2012discovering,pirsiavash2012detecting} & \checkmark &  & \checkmark &  &  &  & & \\
\cite{Sharghi2016}&S& & &&\checkmark&F&\checkmark&	 & &\checkmark	&\cite{lee2012discovering, yeung2014videoset}& & & & & &\checkmark &&\\
\cite{8099601}&S& & & &\checkmark&	F&\checkmark	& & &\checkmark&\cite{lee2012discovering,yeung2014videoset}& & & &\checkmark& &\checkmark &&\\
\cite{7457669}&	S& & & &\checkmark&CA&\checkmark& & & & & &	\checkmark& & & & &&\\
\cite{7422088}&S& & &	\checkmark& &E & & & & &	\cite{7422088} & &\checkmark& & & & &&\\
\cite{sharghi2017query}&	S&& &&\checkmark&F&\checkmark&&&			\checkmark&	\cite{lee2012discovering}&&&&&&\checkmark	&&\\
\cite{xu2008novel} & S &  &  &  &  &  & \checkmark & \checkmark &  & \checkmark &  &  &  &  &  &  & \checkmark & & \\
\cite{7547309}&	{S}& & &\checkmark&	&F&	\checkmark& & & &\cite{gygli2014creating, song2015tvsum, 6126279, 5995586}&& &\checkmark	&\checkmark	&\checkmark	&\checkmark&&\\	
\cite{gao2017event}&S+Mv&&\checkmark &&&G&\checkmark&&&\checkmark&&\checkmark&\checkmark&&&&&&\\
\cite{peng2009user} & Af &  & \checkmark &  &  & CD & \checkmark &  &  & \checkmark &  & \checkmark & & &  &  &  &  &  \\
\cite{joho2009exploiting} & Af &  &  &  &  &  &  &  &  & \checkmark &  & \checkmark &  &  &  &  &  & & \\
\cite{peng2010real} & Af &  &  &  &  &  &  &  &  & \checkmark &  & \checkmark &  &  &  &  & \checkmark &&  \\
\cite{money2010elvis} & Af &  &  &  &  &  &  &  &  & \checkmark &  & \checkmark &  &  &  &  &  &  &\\
\cite{katti2011affective} & Af &  &  &  & \checkmark & CA+E &  &  &  & \checkmark &  & \checkmark &  &  &  &  & & &  \\
\cite{peng2011editing} & Af &  &  &  &  &  &  &  &  & \checkmark &  & \checkmark &  &  &  &  & \checkmark &  &\\
\cite{xiang2011affect} & Af &  & \checkmark &  &  & CD & \checkmark & \checkmark &  &  &  & \checkmark &  &  & \checkmark &  &&  &  \\
\cite{Tang:2012:ECS:2207676.2208622} & Af &  &  &  &  &  &  &  &  & \checkmark &  & \checkmark &  &  &  &  &&  &  \\
\cite{yoshitaka2012personalized} & Af &  &  &  &  &  &  &  &  & \checkmark &  & \checkmark &  &  &  &  &  &  &\\
\cite{xu2015gaze} & Af &  &  &  & \checkmark & E & \checkmark &  &  & \checkmark & \cite{xu2015gaze,fathi2012learning} & \checkmark &  &  & \checkmark &  &  &  &\\
\cite{ma2005generic} & At & \checkmark &  &  &  &  & \checkmark & \checkmark &  &  &  & \checkmark &  &  &  &  &  &  &\\
\cite{ngo2005video} & At &  & \checkmark & \checkmark &  & MT+G & \checkmark &  &  &  &  &  & \checkmark & \checkmark &  &  &  &&  \\
\cite{you2007multiple} & At & \checkmark & \checkmark & \checkmark &  & ID & \checkmark &  &  &  &  & \checkmark &  &  &  &  &  &&  \\
\cite{evangelopoulos2008movie} & At & \checkmark &  &  &  &  & \checkmark & \checkmark &  &  &  & \checkmark &  &  &  &  &  &  &\\
\cite{evangelopoulos2009video} & At & \checkmark &  &  &  &  & \checkmark & \checkmark & \checkmark &  &  & \checkmark &  &  &  &  &  &  &\\
\cite{evangelopoulos2013multimodal} & At & \checkmark &  &  &  &  & \checkmark & \checkmark & \checkmark &  & \cite{spachos2008muscle} & \checkmark &  &  &  &  &  &  &\\
\cite{7351630} & At & \checkmark &  &  &  &  & \checkmark & \checkmark & \checkmark &  & \cite{7148146} &  &  &  &  &  & \checkmark &  &\\
\cite{mehmood2016divide} & At+Af & \checkmark & \checkmark &  &  & CD & \checkmark & \checkmark &  & \checkmark &  & \checkmark &  &  &  &  &  &  &\\
\cite{7362797} & At+Af & \checkmark &  &  &  &  &  & \checkmark & \checkmark &  & \cite{7148146} & \checkmark &  && & & \checkmark &&  \\
\cite{ren2010semantic} & S+At &  &  &  & \checkmark & E & \checkmark & \checkmark & \checkmark & \checkmark &  & \checkmark &  &  &  &  &  &  &\\
\cite{han2011personalized} & S+At & \checkmark &  &  &  &  & \checkmark &  &  &  &  &  & \checkmark &  & \checkmark &  &  &&  \\
\cite{yuan2011video} & S+At &  &  &  & \checkmark & ID & \checkmark &  &  &  &  &  &  &  & \checkmark &  &  &  &\\
\cite{dong2014iteratively} & S+At &  &  &  & \checkmark & F & \checkmark &  & \checkmark &  &  & \checkmark &  &  &  &  &  &  &\\
\cite{7931584}&S+At& & & &\checkmark&	L&\checkmark	& & &\checkmark&\cite{7931584}& & &\checkmark&	\checkmark& &\checkmark&&\\
\cite{Varini:2015:EVS:2733373.2806367}&	S+At	&	\checkmark	& & & &	&	\checkmark&&&	\checkmark&		&	\checkmark	&&&&&	\checkmark	& &\\	
\cite{xu2016fast}&S+Af&	&&&\checkmark&	F&	\checkmark&&&&	\cite{Jiang:2011:CVU:1991996.1992025}&	\checkmark&&&&&&&\\
\cite{7904630}&ML& & & &\checkmark	&F&\checkmark& & & &\cite{gygli2014creating, song2015tvsum,pirsiavash2012detecting}& & & &\checkmark& &\checkmark&&\\
\cite{khosla2013large}&ML&\checkmark& & & & &	\checkmark	&&&		\checkmark&				&&&&\checkmark	&	&\checkmark&	&\\
\cite{potapov2014category} & ML &  &  &  & \checkmark & MT & \checkmark &  &  &  & \cite{potapov2014category} &  &  &  &  &  & \checkmark & & \\
\cite{gygli2015video} & ML &  &  &  & \checkmark & F & \checkmark &  &  &  & \cite{lee2012discovering,gygli2014creating} &  &  &  & \checkmark &  &\checkmark  &&  \\
\cite{8099801}&ML&\checkmark & & &	 & &\checkmark& & & &\cite{gygli2014creating, song2015tvsum,OVproject} & & & & & &&&\checkmark \\
\cite{panda2017weakly}&ML&&\checkmark&&&ID&\checkmark&&&&\cite{chu2015video,song2015tvsum}&&&&&&&&\checkmark\\
\cite{sun2014ranking}&ML&&&&\checkmark	&F&	\checkmark&&&&\cite{sun2014ranking}&&&&&&\checkmark&&\\
\cite{yang2015unsupervised}&ML&	&&&\checkmark&	MT&\checkmark&&&&	\cite{sun2014ranking}&&&&&&&&\checkmark\\
\cite{gygli2016video2gif}&ML&&\checkmark&&&	MT&\checkmark&&&&	\cite{sun2014ranking,gygli2016video2gif}&&&&&&\checkmark&&\\
\cite{yaohighlight} & ML &  &  &  & \checkmark & MT & \checkmark &  &  &  &  &  &  &  &  &  &  & &\checkmark \\
\cite{zhang2016summary} & ML &  &  &  & \checkmark & MT & \checkmark &  &  &  &  \cite{gygli2014creating,potapov2014category,Avila} &  &  &  & \checkmark &  & \checkmark &&  \\
\cite{zhang2016video} & ML &  &  &  & \checkmark & MT & \checkmark &  &  & & \cite{gygli2014creating,song2015tvsum,OVproject,Avila} &  &  &  &\checkmark &  &  & &\checkmark \\ 
\cite{zhao2017hierarchical}&ML&&&&\checkmark&F&	\checkmark&	&&&			\cite{zeng2016generation,gygli2014creating,song2015tvsum,potapov2014category}&&&&&&&&\checkmark\\
\cite{ho2018summarizing}&ML&&&&\checkmark&F&	\checkmark&&&&&&&&&&	&&\checkmark\\
\cite{wei2018video}&ML&&\checkmark&&\checkmark&MT&\checkmark&&&\checkmark&\cite{gygli2014creating, song2015tvsum}&&&&&&&&\checkmark\\
\cite{lin2015summarizing}&ML+O&&&&\checkmark&	F&	\checkmark&&&&	\cite{sun2014ranking}&&&&&&\checkmark&&\\
\cite{ji2017video}&ML+At&&&&\checkmark&MT&\checkmark&&&&\cite{gygli2014creating, song2015tvsum}&&&&&&&&\checkmark\\
\hline
\end{longtable}
\end{center}

\section{Evaluation and datasets}
\label{sec:evaluation}
This section gives a brief overview of the different evaluation methods along with the relevant and available datasets for video summarization. A video skim is a consumable for human viewers, so it would be justified if subjective evaluation of the video skim is performed. Earlier video skimming techniques extensively relied on subjective feedback for assessment. Gradually, datasets with user created ground truth summaries for evaluation emerged. \cite{taskiran2006evaluation} describes intrinsic and extrinsic methods for evaluating video summarization algorithms based on text summarization evaluation. Intrinsic methods rely on the analysis of the summary either subjectively or by comparing with the user generated summary. Whereas extrinsic methods evaluate the summary with respect to an information retrieval task. Both qualitative and quantitative methods are important to assess video skimming algorithms, which will be discussed in Sections \ref{sec:qualitative} and \ref{sec:quantitative} respectively. Section \ref{subandobj} discusses works that evaluate the video skimming performance using both qualitative and quantitative metrics and Section \ref{textbasedevaluation} discusses works that utilize text based evaluation of video skimming performance. Datasets with ground truth for different video types are listed and discussed in Section \ref{datasetsection}. The performance of a few significant video skimming approaches on some of the datasets are discussed here as well.

\subsection{Qualitative Evaluation}
\label{sec:qualitative}
Qualitative evaluation is required due to the subjective nature of the result. The generated summary may please one viewer, whereas, some other viewer may not be satisfied with it. Each viewer has his/her preferences or in other words, expectations from a summary in order to accept it as a `good summary'. Due to the inherent subjectivity of the result, it is necessary to consider human evaluators. In a subjective/qualitative evaluation scheme, users after watching the complete video and its summarized version, give scores for each of the considered criteria. The different criteria used are described below.
\subsubsection{User scoring:}
Here users are asked to rate the video skim with respect to the original video. \cite{smith1998video, taskiran2006automated, gao2009dynamic} use criteria of informativeness, compression ratio and satisfaction, where  informativeness says about the amount of information carried over to the summary, compression ratio refers to the length of summary achieved with respect to the original video, satisfaction refers to the enjoyability of the summary. \cite{xu2008novel} uses criteria based on completeness, smoothness and acceptance. Completeness measures whether the summary includes all the events related to the query, smoothness measures the viewing comfortableness and acceptance measures the percentage of valid summary. \cite{fu2010multi,ngo2005video} use enjoyability, informativeness and usefulness. \cite{tsai2013scene} evaluates the movie summaries using enjoyability, informativeness, plot understanding, major role relationship understanding, story understanding and overall quality. Apart from that the authors took help of professional editors to manually create movie summaries in order to measure the hit rate of the scene length, and role community distribution in automatic and manual summaries. In~\cite{chen2009novel} evaluation is based on access to informativeness and interrelation of the video skim.  \cite{lu2013story, xiang2011affect, katti2011affective} measures the percentage of users preferring the summary. In \cite{ma2002user,you2007multiple,evangelopoulos2008movie,evangelopoulos2009video,evangelopoulos2013multimodal} subjects are given the privilege to choose the skim ratio. They use criteria of informativeness and enjoyability to assess the skims. Here enjoyability accesses the smoothness of the summary in terms of speech comprehensibility. \cite{darabi2015personalized} does a user study based on four metrics: precision, recall, timing and overall satisfaction. ``Recall measures the capability of the system in the terms of coverage of the whole video, or in other words, the extent to which the generated summary can reflect all the scenes from the original video. Precision measures the ability of the generated summaries to include the most important scenes of the original video. Timing explains the level of temporal proximity of the built abstracts to the required summary length and `overall satisfaction' score represents the extent to which the end users are satisfied with the summaries from different points of views, namely visual and aural coherency, continuity and adjustability.''  In \cite{Chen2003358}, the criteria of coverage, quality and interestingness are used. In \cite{7931584}, user rating is collected in terms of fidelity of user preferences included in the summary and the visual smoothness of the generated summary. \cite{xu2016fast} performs subjective evaluation in terms of accuracy, emotion, coverage and quality.

\subsubsection{User scoring based on available summaries:} 
Here users are asked to rate the video skim by comparing with a standard video skim created professionally. \cite{peng2008aesthetics} compares the summary with the ones generated manually using commercially available software like PowerDirector \cite{PowerDirector} and MuVee \cite{muvee}. User feedback is taken using a questionnaire on the satisfaction of the summary and whether the technique can drop underexposed and blurred shots. In \cite{hong2011beyond}, videos are collected from YouTube for specific `event queries'. Video skims are evaluated based on `informativeness', `experience' and `acceptance', where informativeness means coverage, experience measures the usefulness of the summary in understanding the events and acceptance is the willingness of the user to use the summarization scheme. \cite{ren2010semantic} uses official trailers of the movies to identify the inclusion of key shots in the generated summaries, here a human evaluator scores the summary based on how much content is matching between the automatic summary and trailer. \cite{kannan2015you} uses a questionnaire, where the viewers are asked to compare an user-generated summary with an automatic summary, and then give the rating in terms of informativeness, enjoyability, relevance, and acceptance. Here relevance measures the closeness of summary to the given preferences. Quality of perception based on information assimilation and satisfaction criteria is used in \cite{gulliver2006defining}. \cite{peng2010real, peng2011editing}, compare the automatic summary to the one generated by random shot selection and to the summary generated by a novice user, in terms of user satisfaction scores.

\subsection{Quantitative Evaluation}
\label{sec:quantitative}
With the availability of video datasets with annotated user summaries, an objective way of summary evaluation based on `precision' (p), `recall' (r) or `f-measure' (f) is being used. Given a Reference summary (ground truth, R) and a Candidate summary (automatically generated video skim, C), then $p=R \cap C/|C|, r=R \cap C/|R|$, and $ f=(2*p*r)/(p+r)$, where $|\cdot|$ implies number of frames and $\cap$ implies common frames. As the datasets consist of several reference summaries per video, the f-measure is calculated with respect to each of the reference summaries and the mean over all the reference summaries is considered for the video. Average of the mean f-measures (Avg. f-measure) for all the videos in a dataset is considered as the performance on that dataset. Some works select maximum f-measure among the reference summaries and consider the average over all the videos as a performance measure (Avg. max. f-measure). Works using a dataset consisting of a collection of videos on specific topics have usually employed mean average precision (mAP), i.e., ``mean of average precision over all categories'' \cite{chu2015video} to measure video category wise performance.

Earlier, techniques used custom man-made reference summaries for comparing the video skims. Eventually datasets were created for the purpose of benchmarking. We follow the same pattern and provide an overview of the techniques that created custom reference summaries followed by representative datasets available in the field.

\subsubsection{Using custom human-made summary:}
\cite{dong2010video,dang2014heterogeneity,zhao2014quasi,chu2015video,han2011personalized,dong2014iteratively,peng2009user,joho2009exploiting,money2010elvis, yoshitaka2012personalized,mehmood2016divide,7952384} created self-compiled ground truth for the collection of videos and evaluate the summary by calculating the inclusion rate (frame-wise recall) between automatic and user generated summary. \cite{khosla2013large,6909934} compares automatic summaries with several user created summaries (utilizing Amazon MT). Equal number of frames are extracted for automatic summary as in the reference summary. The pixel wise difference between the ShiftFlow \cite{5551153} transform of summary frame and reference frame is used to find the best match. Precision is found out based on the number of matching frames. In \cite{7457669}, entity based summarization is performed. The evaluation is done in terms of  ``conciseness, defined as the ratio of entity coverage to the number of significant clusters, and representativeness, defined as the ratio of the tracklet coverage to the number of significant clusters'' \cite{7457669}.

\subsubsection{Using dataset:} The various benchmarked datasets used by respective articles are shown in Table \ref{tableAV}. The descriptions of these datasets are given in Section \ref{datasetsection}.  As a video in a dataset consists of multiple ground truth summaries, Avg. f-measure is used for comparing the performance in general. Avg. max. f-measure is employed in \cite{zhang2016video,8099801}. mAP is used in \cite{chu2015video,8099938,lin2015summarizing,yang2015unsupervised,gygli2016video2gif,sun2014ranking,panda2017weakly}. Importance ratio and meaningful summary duration are employed for evaluation in \cite{potapov2014category}. Importance ratio is the ratio between the scores achieved by the summary to the maximum score that can be achieved by selecting segments using a greedy approach. Meaningful summary duration (MSD) measures the summary duration required to capture the gist of the video. Lower the summary duration better is the summarization technique. As MSD is influenced by the length of the ground truth with respect to the video, \cite{gygli2016video2gif} suggests using normalized MSD. In \cite{7780487} evaluation is done using recall, which is calculated using the number of desired objects in reference and candidate summary. Average recall over all the object queries is also considered as a measure of performance over a dataset \cite{4509438}.  In \cite{8099680}, the summaries are not constrained by summary length, so the traditional f-measure is not suitable. Instead they use Mathews Correlation coefficient \cite{matthews1975comparison} for evaluating performance. This coefficient takes into consideration the specificity (true negative rate) during evaluation and provides a score in (-1,1), where -1 means perfect disagreement with ground truth and 1 means perfect agreement.

\subsubsection{TRECVID rushes challenge:}
TRECVID \cite{over2007trecvid,over2008trecvid} challenge has nine criteria: 3 objective, 2 usability and 4 subjective measurements. They are: ``duration of summary in seconds, difference between target and actual summary size, fraction of inclusions found in the summary, degree of junk frames in summary, degree of duplicate video in summary, degree of pleasant tempo/ rhythm in the summary, total time spent in judging the inclusions in seconds, total video play time judging inclusion, total running time in seconds'' \cite{over2007trecvid}. These criteria are used by \cite{4907069,Valdes:2008:BTB:1463563.1463588,5993544,laganiere2008video,almeida2013online}.

\subsection{Both Subjective and Objective Evaluation}
\label{subandobj}
In \cite{7368910,6671471,7121011,gao2017event} and \cite{Wang:2014:RSU:2647868.2655013} perform subjective evaluation along with objective evaluation using datasets. \cite{7121011} uses the subjective criteria of pleasantness and visual informativeness to evaluate multi-view summary. In \cite{Wang:2014:RSU:2647868.2655013}, criteria of accuracy and coverage are used. ``Accuracy measures the relevance of a skim to the dominant high-level semantics of corresponding original video, and coverage verifies whether a summary contains sufficient information to understand the original story with little content redundancy and quality'' \cite{Wang:2014:RSU:2647868.2655013}. 

\subsection{Text based Evaluation}
\label{textbasedevaluation}
\cite{yeung2014videoset} suggests using the text annotations of the videos to evaluate the video summary. Here comparison is done between the annotation of the video and its summary using existing text summary evaluation package \cite{lin2004rouge}. \cite{gygli2015video,Sharghi2016,8099601} adopted text based evaluation, along with it a new metric called hitting-recall is suggested in \cite{Sharghi2016}, which is the ratio between the number of query relevant shots in the summary to that in the ground truth. \cite{sharghi2017query} uses f-measure calculated on the basis of the common semantic concepts present in the text description of the summaries.

\subsection{Datasets for Evaluation}
\label{datasetsection}
 The datasets used for the evaluation of dynamic summary generation are listed in Table~\ref{datasettable} along with their descriptions. There are a few other datasets used primarily for static (key frame) based video summarization, which can be adapted for video skimming evaluation by annotating them with important video segments that are essential to be included in video skim. One such dataset is the TRECVID dataset \cite{trecvidevaluations} consisting of news and sports videos. The Open Video Project \cite{OVproject} is a digital video library containing several videos among which some have been used for static video summarization by creating ground truth key frames \cite{Avila} in \cite{Avila,Furini:2010:SSM:1713230.1713242}. The Kodak dataset \cite{loui2007kodak} consists of around 3000 consumer videos. These videos are annotated with semantic concepts at a key-frame level which can be used for video skimming evaluation.

Among the datasets created for video skimming, we are currently witnessing an increase in the creation of datasets for user videos, mainly due to the availability of video websites like YouTube etc. The most popular are TVSum~\cite{song2015tvsum} and SumMe~\cite{gygli2014creating}, as these are the first set of datasets to appear for video skimming along with ground truth. The YouTube Highlight~\cite{sun2014ranking} and MED~\cite{potapov2014category} consists of categorically organized videos, which are helpful for category based learning techniques and so is the GIF dataset~\cite{gygli2016video2gif} that consists of thousands of videos along with user created highlight. Personalization can be achieved from Personalized Highlight dataset~\cite{del2018phd} that includes user identifier into the video highlights created by them.

User videos are of shorter duration, typically of 5 to 10 minutes, whereas egocentric and movies are of longer duration. The prominent dataset for egocentric videos is UTE~\cite{lee2012discovering}, that has been annotated by a few authors to adapt for text based evaluation. The dataset is densely annotated with concepts which helps in semantic processing. The ADL~\cite{pirsiavash2012detecting} and Disney world~\cite{6247805} datasets consists of egocentric videos which can be combined with some adaptation to suit the needs. The Cognimuse~\cite{zlatintsi2017cognimuse} dataset is available for movie summarization, which has ground truth summaries for each of the movies along with saliency and emotion annotations. Saliency and emotion models can benefit from such a dataset. Generally, creating a dataset for long videos with annotations may not be feasible. In such cases, multiple views with annotations   from surveillance or sports datasets can be combined to generate a long video with annotations. It will be challenging to witness the performance of mono-view video summarization techniques on such long videos.

Further, datasets augmented with additional information like Human eye gaze (Egosum+gaze\cite{xu2015gaze} and GTEA gaze+~ \cite{fathi2012learning}), GPS information (Art city~\cite{7931584}) can be helpful for exploring the summarization efforts from novel perspectives.

\begin{center}
\tiny
\begin{longtable}{|p{.20\textwidth}|p{.10\textwidth}|p{.60\textwidth}|}
\caption{Datasets for evaluation of video skim categorized on the basis of video domain. }
\label{datasettable} \tabularnewline
\hline
\textbf{Category} & \textbf{Dataset} & \textbf{Description} \\
\hline
\endfirsthead
\multicolumn{3}{c}{{\bfseries \tablename \thetable{}-- continued from previous page}}\\
\hline
\textbf{Category} & \textbf{Dataset} & \textbf{Description} \\ 
\hline
\endhead

\hline \multicolumn{3}{|r|}{{Continued on next page}}\\ \hline
\endfoot

\hline 
\endlastfoot

\multirow{7}{*}{\textbf{Egocentric}} & UTE \cite{lee2012discovering} & Contains 4 videos recorded using wearable cameras that log the activities of the wearer. Each video is of 3 to 5 hours duration and totals to 17 hours of egocentric video. The dataset is annotated in \cite{yeung2014videoset} using text, as a result text-based summary evaluation is feasible. This dataset is annotated densely with 48 semantic concepts in \cite{sharghi2017query} using Amazon MTurk (AMT). \\ \cline{2-3} 
 & ADL \cite{pirsiavash2012detecting} & ``Contains 20 videos from chest mounted cameras each of about 20 to 60 minutes duration'' \cite{lu2013story}. The videos are annotated with a set of 42 objects. \\ \cline{2-3} 
 & Egosum +gaze\cite{xu2015gaze} & Contains egocentric data acquired by five subjects, wearing an eye-tracking device for collection of gaze data. It consists of 21 videos each lasting 15 minutes to 1.5 hours. This dataset provides human-generated summaries along with annotations. \\ \cline{2-3}
 
  &Disney world \cite{6247805} &This dataset consists of 8 egocentric videos, each of 6-8 hours duration, recorded using a GoPro wearable camera at 30 frames per second with a resolution of 1280 x 720 pixels. Each video records subjects day during a visit to Disney World Park. Three videos are provided with text annotations and ground-truth summaries.\\ \cline{2-3}
 
  &Art City \cite{7931584}&	This dataset consists of 48 videos captured by tourists with a head-mounted camera covering the heritage sites in Italian cities. Video length range from 2 minutes to 15 minutes with resolution in the range $720\times576$ to $1920\times1080$. ``Annotations are added on three different semantic dimensions: observer's behavior, presence of points or items of interest and narrativity related'' \cite{7931584} information.  The videos have GPS annotations with a granularity of 1 second in time and 1 meter linear displacement in space.\\  \cline{2-3}
  
  &CSumm \cite{del2017active} & Videos are recorded using Google Glass at 29 frames per second, with resolution of $720\times1280$ pixels. The length ranges from 15 to 30 minutes. The ``videos include a large selection of activities, such as practicing or watching sports, enjoying nature, having dinner, etc.'' \cite{del2017active}. \\ \cline{2-3}
  
  &UnSum  \cite{ho2018summarizing} &  A collection of 70 first-person videos, including GoPro sports videos from YouTube as well as videos from existing HUJI \cite{poleg2016compact} video-indexing dataset. Among them five videos are provided with shot-level annotations.\\

  \hline
 
\multirow{2}{*}{\textbf{Action/Affective}} & GTEA-gaze+ \cite{fathi2012learning} & It ``is designed for action recognition but can be used for summarization purpose'' \cite{xu2015gaze}. It consists of 30 cooking videos, and each video lasts for 12 to 20 minutes. Action annotations are available with the dataset which can be used for summarization purposes as mentioned in Section \ref{textbasedevaluation} \\ \cline{2-3} 
 & \cite{baveye2015liris} & A video dataset for affective content analysis. Consists of 9000+ video clips with annotated emotions. \\ \hline

\multirow{2}{*}{\textbf{Movie} }& Cognimuse \cite{spachos2008muscle,7148146,zlatintsi2017cognimuse} & Annotated multi-modal movie dataset. Dialog and saliency annotations are available for audio, video, audio-video and text. \\ \cline{2-3} 
& \cite{yeung2014videoset} & This dataset consists of Four TV episodes (1 from Castle, 1 from The Mentalist, and 2 from Numb3rs) of 45 minutes each. The dataset has text annotations and ground truth summaries.\\

  \hline
\textbf{Rushes} & \cite{over2007trecvid,over2008trecvid} & BBC rushes video summarization and evaluation, a workshop organized by TRECVID. Consists of a collection of unedited videos of some BBC drama programs. 42 videos were provided to participants for use in developing their algorithms and 40 were withheld for testing purpose. Duration of the videos is in the range of 9.8 to 36.79 minutes. Ground truth summaries are available for all the videos. \\ \hline
\multirow{2}{*}{\textbf{Sports}} & \cite{7368910} & A collection of tennis videos from 2013 and 2014 Roland Garros tournament. Consists of 12 matches with total duration being more than 28 hours. It is accompanied with editorial summaries which serve as ground truth. \\ \cline{2-3} 
 & \cite{6422365} & A collection of  ``10 soccer videos which are gathered from several countries and broadcasters'' \cite{6422365} with a total duration of 9 hours. The videos are recorded at 25 frames per second with a resolution of $640\times368$ pixels. Videos are annotated with shot boundaries and play-break sequences; a total of 3452 shots with an average shot length of 9 seconds. The dataset is also annotated with soccer events like goal, card, goal attempt, etc.\\ \hline 
\multirow{7}{*}{\textbf{Surveillance}} & `BL-7F' \cite{6838965} & A collection of videos recorded by 19 surveillance cameras. The ``density of the surveillance system is high in order to simulate the scenario of wireless video sensor networks with many overlapped fields of views. All the cameras are fixed and perfectly synchronized'' \cite{6838965}. Important events are annotated.\\ \cline{2-3} 
 & \cite{fu2010multi} & Consists of a collection of multi-view videos from an office environment. ``Lengths of three views of the office videos are 180 minutes 41 seconds, 170 minutes 46 seconds and 176 minutes 43 seconds separately. All of the videos are captured using the web cameras or handheld video cameras by non-specialists, making some of them unstable and obscure. Moreover, some videos have quite different brightness across multi-views'' \cite{fu2010multi}. Important events are annotated.\\ \cline{2-3} 
 & \cite{blunsden2010behave} &  ``Comprises of two views of various scenario's of people acting out various interactions. The data is captured at 25 frames per second. The resolution is 640x480. The videos are available either as AVI's or as a numbered set of JPEG single image files. Most of the video sequences have ground truth bounding boxes of the pedestrians in the scene'' \cite{blunsden2010behave}. \\ \cline{2-3} 
 & \cite{jiastdataset} & A collection of video data used for people tracking, abnormality detection and report generation. The videos are recorded at 30 frames per second in a rectangle lab room, where the action zone is around 3.5m $\times$ 4.5m. \\  \cline{2-3} 
 &UCLA \cite{6126279}& ``It consists of three surveillance videos of single and two-person activities. The total length of these three video sequences is around 35 minutes. Every video sequence is composed of repetitive events with different temporal orders''  \cite{7547309}.\\\cline{2-3} 
 &VIRAT \cite{5995586}& ``There are 334 videos, each lasting 2 to 15 minutes. These videos are recorded on 10 different scenarios, including parking lots, university campuses, etc.''  \cite{7547309}.\\ \cline{2-3}
 
& \cite{7422088}&  A collection of ``25 real traffic surveillance videos from publicly accessible online web cameras in Budapest, Hungary. These videos are combined with two surveillance video data sets, Junction and Roundabout \cite{li2012learning} for a total of 27 videos'' \cite{7422088}. \\  \hline
\multirow{9}{*}{\textbf{User/Home}} & SumMe \cite{gygli2014creating} & Consists of 25 user videos belonging to different categories like a holiday, sports and events. Video duration ranges from 1 to 6 minutes. Each video is annotated with scores by at-least 15 viewers. This annotation will help in generating user summaries, which can be used as ground truth. \\ \cline{2-3} 
 & TVSum \cite{song2015tvsum} & This dataset consists of 50 videos collected from YouTube, related to 10 different categories along with shot based scores for video segments achieved through crowdsourcing. The scores are based on the relevance of the shot/segment to the video title. \\ \cline{2-3} 
 & Columbia consumer video \cite{Jiang:2011:CVU:1991996.1992025}  & Consists of 9317 web videos that include semantics belonging to events like `baseball', `parade', `beach', `cat' etc. Average duration is 80 seconds and the total duration is 210 hours. Semantic annotations for 20 categories, is done by crowdsourcing using AMT. This dataset is also used in the TRECVID event detection task. \\ \cline{2-3} 
 & MED \cite{potapov2014category} & A Multi-media Event Detection dataset a subset of TRECVID'11, it consists of 160 videos belonging to 10 different events and is well annotated and scored. \\  \cline{2-3}
&CoSum \cite{chu2015video} &	This dataset consists of user videos relating to 10 different topics (``Base jumping, Bike polo, Eiffel Tower, Excavator river crossing, Kids playing in leaves, MLB, NFL, Notre Dame Cathedral, Surfing and Statue of liberty'' \cite{chu2015video}) downloaded from YouTube. Each topic has a set of videos whose total length is in the range 10 to 25 minutes. A total of 51 videos amounting to 147 minutes are available.\\  \cline{2-3}
& YouTube Highlight \cite{sun2014ranking}& A collection of videos belonging to 6 categories namely skating, surfing, skiing, gymnastics, parkour, and dog activity. For each category, there are about 100 videos of varying length totaling to 1430 minutes of video data. The videos in each category are accompanied by edited videos in the category, which can be used for learning. Raw videos annotations are performed using AMT, a single highlight of about 5 seconds is selected as ground truth.\\  \cline{2-3}
& GIF  \cite{gygli2016video2gif} & Consists of 121,647 GIFs with an average duration of 5.8 seconds created from 84,754 videos with an average duration of video being 5.2 minutes. The videos belong to a variety of categories.\\ \cline{2-3}
&Videos in the Wild (VTW)  \cite{zeng2016generation} & 	A collection of 18100 videos with an average duration of 1.5 minutes is crawled from YouTube and used for video captioning purpose. About 2000 videos are labeled with sub-shot level scores and can be used for video summarization purpose.\\ \cline{2-3}
&Personalized highlight (PHD) \cite{del2018phd} &  This dataset is collected from gifs.com, a web  platform for editing videos and creating GIFs. It consists of 13, 822 users with 222, 015 annotations on 119, 938 videos. Personalization is achieved by utilizing the user information attached to the videos, in terms of the segments selected by the user as a highlight. \\ 
\hline
\end{longtable}
\end{center}

\subsubsection{Comparison of Performance}
We collect the results of various video summarization techniques published in the literature to give a comparative overview between them. Among the multiple objective evaluation criteria, we aggregate the results in Table \ref{empiresults} based on Avg. f-measure and mAP as they are the most used ones. Avg. f-measure and mAP are the averaged values of mean f-measure and AP calculated over all the videos and corresponding reference summaries available in a dataset. The best performing techniques are shown in bold. We provide the upper bound of the f-measure metric for some of the datasets for verifying the proximity of the best results to the ideal skim. To do so, we first created oracle summary \cite{gong2014diverse} for each of the videos in the dataset by utilizing the reference summaries. Oracle summary is formed by selecting the frames in such a way that it maximizes the required metric. Second, we restrict the length of the oracle summary to 15\% of the video length, as this is the length preferred in most of the techniques. Third, we assume that the oracle summary is generated by some skimming technique (as this is the best possible skim that can be generated) and evaluate it with respect to the reference summaries in the dataset. We find that the maximum achievable Avg. f-measure for SumMe and MED datasets are 0.492 and 0.654, respectively. This method of finding maximum f-measure cannot be applied for TVSum dataset as in this case, the 15\% reference summary for a given video varied depending on the candidate summary generated by the technique that is being evaluated. In the case of CoSum and YouTube Highlight datasets, oracle summaries are irrelevant. This is because, CoSum dataset suggests the use of an aggregated reference summary derived from three reference summaries for each video. A shot selected at least in two out of the three summaries is considered a part of the aggregated reference summary. A similar suggestion is associated with YouTube Highlight dataset, where the parts of a video forming a single reference summary are obtained through crowdsourcing. As the comparison would be with a single reference summary in both the above datasets, we can conclude that the maximum mAP of 1.0 can be achieved on these datasets.

From the analysis of results, we can conclude that the evaluation datasets are still challenging and there is scope for improvement. For Avg. f-measure based evaluation, the technique of \cite{zhao2017hierarchical} performs best on SumMe, TVSum and MED datasets. The work uses RNN in a hierarchical setting to exploit long-range temporal dependencies among the video. Such a modeling of temporal dependencies accurately captures the redundancies in a video, resulting in the superior performance observed. Among mAP based evaluation, \cite{gygli2015video} performs best on TVSum and CoSum datasets, and \cite{lin2015summarizing} performs best on YouTube Highlight dataset. \cite{gygli2015video} uses sub-modular functions to weight different summarization criteria like uniformity, representativeness, and interestingness, which suggests that the variable weights used for fusing different criteria has resulted in better performance. \cite{lin2015summarizing} uses a structured SVM to learn the difference between highlight and non-highlight segments by considering their context. We think that the use of context for highlight detection has given it an edge over the other techniques. Although the techniques have shown better performance using objective criteria, it is to be verified by subjective evaluation on the consumability of the summaries, which is a possible future work in this area (refer Section \ref{pleasingVS}).

Empirical results for multi-view summarization are also available. As there are relatively less number of such techniques, which are already recorded together in \cite{7121011}, we do not consider them here.

\begin{table}[]
\tiny
\caption{Avg. f-measure and mAP based comparison among a few popular and latest techniques for some commonly used datasets. The values are taken either from respective works or other works that have performed comparative study. The two entries for \cite{zhang2016video} represents learning in canonical mode and augmented mode.  The technique that has the best performance on each of the datasets is shown in bold. `-' represents unavailable dataset technique pair. Acronym N.A implies Not Applicable.}
\label{empiresults}
\begin{tabular}{|l|ccc|p{.25\textwidth}|}
\hline
\multirow{2}{*}{\bf{Ref.}} & \multicolumn{3}{|c|}{\bf Avg. f-measure} & \multirow{2}{*}{\hspace{1cm}{\bf Skimming Model}}\\ \cline{2-4}
& \begin{tabular}[l]{@{}c@{}}{SumMe \cite{gygli2014creating}} \\ \tiny{(\bf max. 0.492)}\end{tabular}&\begin{tabular}[l]{@{}c@{}}  {TVSum \cite{song2015tvsum}}\\ \tiny{(\bf max. N.A)} \end{tabular}&\begin{tabular}[l]{@{}c@{}}{MED \cite{potapov2014category}}\\ \tiny{(\bf max. 0.654)}\end{tabular}&  \\ \hline
\cite{gygli2014creating}& 0.234 &-  &-  & weighted fusion   \\
\cite{potapov2014category} &-  &0.471  &-  &video category specific SVM models   \\
\cite{song2015tvsum}&0.266&0.5&-&video title related images as prior\\
\cite{gygli2015video}&0.397&0.568&0.285&sub-modular functions\\
\cite{zhao2014quasi} &0.384&0.477&0.262&sparse coding\\
\cite{khosla2013large}&0.240&0.360&-&video related images as prior\\
 \cite{zhang2016video}&-&\begin{tabular}[l]{@{}c@{}}\{0.542, \\ 0.579\}\end{tabular} &0.293&supervised LSTM\\
 \cite{zhang2016video}&-&\begin{tabular}[l]{@{}c@{}}\{0.547,\\ 0.596\}\end{tabular}&0.296&LSTM with DPP\\
\cite{zhang2016summary}&0.409&0.541&0.297&\\
 \cite{8099801}&-&0.517&-&GAN with DPP\\
\begin{tabular}[l]{@{}c@{}} \cite{8099801}\end{tabular}&-&0.563&-&supervised GAN\\
\cite{zhao2017hierarchical}&\bf 0.443& \bf 0.621&\bf 0.311&hierarchical RNN\\
\cite{7904630}&0.431&0.527&-&weighted fusion\\ \cline{2-4}
&  \multicolumn{3}{|c|}{\bf mAP}&\\ \cline{2-4}
&\begin{tabular}[l]{@{}c@{}} {CoSum \cite{chu2015video}}\\ \tiny{\bf( max. 1.0)} \end{tabular} & \begin{tabular}[l]{@{}c@{}}  {TVSum \cite{song2015tvsum}}\\ \tiny{(\bf max. N.A)} \end{tabular}& \begin{tabular}[c]{@{}c@{}}  YouTube \\ Highlight\\ \cite{sun2014ranking}\\ \tiny{(\bf max. 1.0)} \end{tabular}&\\ \cline{2-4}
\cite{6247852}&0.506 &0.320  &-  &  sparse modeling  \\
\cite{panda2017weakly}&0.736&0.415&-&deep learning model\\
\cite{potapov2014category}&0.686&0.387&-&video category specific models\\
\cite{gong2014diverse}&0.709&0.435&-&DPP\\
\cite{gygli2015video}&\bf0.745&\bf 0.443&-&sub-modular functions\\
\cite{chu2015video}& 0.579&0.342&-&co-summarization of related videos\\
\cite{8099938}&0.677&0.371&-&co-summarization of related videos\\
\cite{zhao2014quasi}&0.527&0.325&0.420&sparse coding\\
\cite{sun2014ranking}&-&-&0.536&ranking SVM model\\
\cite{yang2015unsupervised}&-&-&0.434&recurrent auto encoders\\
\cite{lin2015summarizing}&-&-&\bf 0.550&structured SVM model\\
\cite{gygli2016video2gif}&-&-&0.464&deep learning model\\ \hline
\end{tabular}
\end{table}

\section{Discussion and Concluding Remarks }
\label{subs:scope}
The primary advantage of a video skimming system is that of easy and quick comprehension of the significant events in the video. When this is expanded to multi-view videos, it also helps in reducing the duplication of video segments across perspectives. When applied in real-time and online modes, it helps in reducing storage space required. The application of video skimming has a wide range of benefits, but along with that there are several key challenges for researchers pursuing this field. We shall now discuss some important issues regarding video skimming followed by possible future research directions.

\subsection{Present Challenges}
Here we discuss some challenges that the already existing approaches can look to counter.
\begin{enumerate}
\item Ground truth for deep learning: Advanced learning approaches have become popular tools that are effectively used in many computer vision tasks including video summarization. With the wide availability of GPU's, a wide spectra of deep learning approaches for dynamic summarization could be made possible by creating a large dataset with associated ground truth. With such systems, the availability of large training data will be a bottleneck, which can be overcome by dataset adaptation and by creating synthetic datasets (refer Section \ref{syndataaset}). In addition to this, the computational time for training and memory requirements have to be studied along with the applicability of supervised techniques for online and real-time video skimming. The lack of large training data can also be tackled by devising unsupervised deep learning approaches \cite{8099801}.
\item Lengthy video: Most of the existing techniques focus on summarizing short videos like consumer videos. The use of such techniques can be extended to much longer videos such as movies and surveillance. The challenge regarding long videos is the computation time requirement for performing segmentation followed by selection. A segmentation technique which can also emphasize on the significance of segments will be of greater value. Alternately, choosing frames uniformly and applying the summarization algorithms would result in feasible solutions. In lengthy movies, where a collection of shots form scenes, the segmentation can be done in a hierarchical way, but it will be a time consuming process. The activities in the movie changes as the plot advances, so in order to do away with the analysis of scenes and shots, it might be suitable to segment the movie into skim units of predefined duration and perform summarization. Thus, by doing so the algorithm could be significantly faster without much difference in performance than when hierarchical segmentation is performed.
\item Real time summarization: Although there are a few works on online video processing for summarization, real-time algorithms for video skimming have not got the emphasis from researchers. Segmentation and feature extraction, which takes up the maximum time, needs to be made faster. The video coding technique will also be of help while analyzing videos for summarization. Given the limited amount of available frames, some insights from coding techniques can be used for tackling the task of segmentation. 
\item Video processing to multi-media processing: As seen in Table \ref{tableAV}, many sophisticated approaches work only on single modality, video, to perform summarization. Future approaches need to consider a shift to synergistic multi-modal processing, considering other kinds of media such as audio, text and meta-data, as they all can complement each other. In addition, appropriate fusion techniques for rating the collective effect of available modalities is an open avenue. The current approach of fusing modalities is that of weighted combination. Intelligent fusion schemes need to be designed that can assess the contribution available from each of the modalities towards summary generation, and use it in skimming.
\end{enumerate}

\subsection{Future Directions}
Using the generic framework of Fig. \ref{fig:blockdiagram} more investigations to improve algorithms may proceed in the following directions.

\subsubsection{Summary Length}
A user defined parameter in all of the techniques is the `summary length'. Is there an ideal summary length? Can it be determined from the video length itself? Not much thought has been given to these questions. The default understanding is that the viewer requires a specified length that corresponds to the time he/she is willing to spend on viewing the video. Alternately, there can be videos which cannot be shortened. This discussion is related to an important research gap, that of deciding the optimal summary length of the video. The lack of understanding of an optimal summary length creates an issue during the evaluation of a generated summary. Since user created summaries (ground truth) could be of variable lengths, what would be a proper way of evaluating an automatically generated summary (of user-defined length) using them? A solution could be not to restrict the summary to a particular length. The viewer can have his/her preferred length and choose the viewing time. When such a provision is not available, only critical segments can be shown to the viewer. Critical segments are those parts of the video that have to be present in the summary. Irrespective of the preferred length, all the critical segments should be included in the summary. In case the preferred length being greater than sum of lengths of critical segments then other segments could be selected into the summary according to certain rankings.

\subsubsection{Pleasing Video Skims}
\label{pleasingVS}
As pointed out in \cite{PFEIFFER1996345}, there needs to be a restriction on the minimum duration of a video skim unit (generated by segmentation), so that the generated summaries are comprehensible and pleasant to view. It was suggested that a suitable viewing chunk of at least about 3.5 seconds would be necessary. However, such a restriction could be an untenable constraint for an algorithm. Hence, research is necessary in a seamless combination of successive important segments to generate the summary. This would be a deviation from the current approach of putting segments together sequentially to get the skim.

\subsubsection{An Adaptive Framework}
The current techniques focus on specific video domains and choosing appropriate summarization criteria according to it. It would be helpful, if a generalized approach is designed that can adapt to different video domains producing suitable dynamic summary. As an alternative, a pre-processing block can be added to the framework, possibly signifying the video domain.

If such domain adaptation is considered, then the video can be viewed as a set of activities; activity refers to any action or movement made by either the objects in the scene or by the camera. Among those activities, the summarization algorithm can rate the impact of an activity relative to all activities in the video. As this operation performs a kind of generic activity detection, it can help in creating generic video skimming algorithms. We believe the generic activity detection is quite feasible, as it is along the lines of generic object detection, which has already been achieved \cite{redmon2016you}.

\begin{figure}[t]
\centering
  \includegraphics[height=4cm,width=9cm]{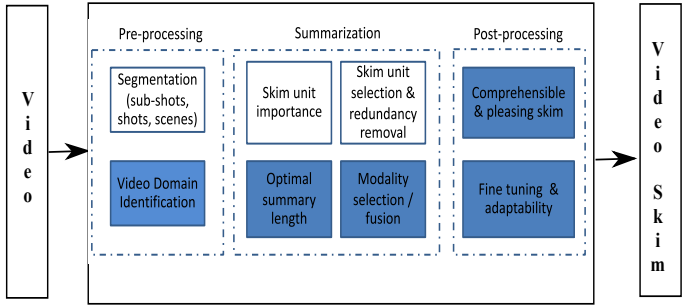}
  \caption{Video skimming system structure in near future (Blue blocks represent possible future advancements) }
\label{fig:enhancedblockdiagram}
\end{figure}

In addition, if required, the framework should  be capable of generating both static and dynamic summaries. An extended framework based on the aforesaid future directions is shown in Figure \ref{fig:enhancedblockdiagram}, where the highlighted blocks are the ones related to future directions.

\subsubsection{Synthetic Dataset}
\label{syndataaset}

An approach for solving the dataset problem (limited amount of data, usually consisting of few videos) is that of `dataset adaptation'; it is a task of adapting the existing datasets for other computer vision tasks to the requirements of video skimming. This significantly reduces the effort involved in preparing the dataset from scratch and is a good strategy for generating a large amount of videos with ground truth. Such methods have been utilized for object recognition tasks \cite{Saenko:2010:AVC:1888089.1888106}.  Another way of generating large dataset is to learn the rules for object interactions from real videos, to create a synthetic video. A synthetic video may consist of moving structures similar to animated videos. This method is analogous to statistical bootstrapping, where using few available samples, large amount of sample are generated. Unlike videos capturing the natural scene, summarizing graphical objects and structures will be an interesting topic to study.

\subsubsection{Evaluation}
Evaluation can be qualitative as well as quantitative. In spite of the developments in quantitative evaluation, equal focus should be given to each of these, as the viewer remains the best judge to approve the summary. It is desired to devise ways to quantify subjective criteria like satisfaction, similarity, enjoyability, completeness, etc. Therefore a qualitative evaluation agreeing with quantitative evaluation will provide for a multi-attribute based assessment of a summary, which can instigate research towards designing novel paradigms for video skimming. 

Quantitative evaluation is performed using precision and recall through f-measure. Since video summarization is about reducing the video length, precision should be given more emphasis because it determines whether the automatic summary is containing prominent amount of ground truth. Although, mAP does emphasize precision but ignores recall entirely. With precision-recall calculation, there is a kind of matching between the frames of automatic and reference summary. For video skimming, an alternate matching mechanism would be to match the frames not only when there is an exact match but also when the context is similar. For e.g., the selection of ${(i+1)}^{th}$ frame in automatic summary instead of $i^{th}$ frame should not be penalized as long as the context between the two frames has not changed. Further, when multiple ground truth summaries are available for a video, the mean f-measure is considered. This method ignores how the ground truth summaries agrees with each other, which should be considered while combining the f-measures.

\subsection{Conclusion}
\label{sec:conclusion}
We have provided a literature survey on dynamic video summarization techniques with specific emphasis on the last decade. A generic framework and taxonomy for dynamic summarization is also proposed. The evolution of techniques from conventional, which use low level features, to the sophisticated approaches that focus on generating personalized summaries based on analyzing human aspects are provided in detail, including those involving latest deep learning paradigms. We find that among the human aspects that have been considered, a hierarchy needs to be defined among them from the viewers' point of view, which may help is designing a skimming approach based on their fusion. We also note that most of the video skimming algorithms are viewer centric and as every viewer has different preferences, it is difficult to find an ideal summary. While there is a need for novel summarization paradigms, we think that the intentions of the videographer has to be represented in the summary more than that of the viewer. The viewers can have varied preferences, but the videographer might have unique objectives.

A detailed description of the available datasets for video skimming from various video domains is also provided. The subjective and objective criteria used for the evaluation of automatic video summaries are also discussed. We have suggested the requirement for conducting video skimming challenge for the purpose of benchmarking the computational efficiency of the algorithm as a part of the evaluation. Techniques such as dataset adaptation and synthetic datasets for creating larger datasets for massive machine learning as well as experimenting video skimming on animated/synthetic videos were also suggested.

We finish by elaborating on the present challenges and highlighting possible future directions, which include real time video summarization, applying video skimming algorithms on longer videos, generic skimming, etc., for research in dynamic video summarization. 

\section*{Acknowledgements}
This work is supported by the Ministry of Human Resource Development (MHRD), Government of India under grant OH-31-23-200-428.

The authors would like to thank all the reviewers' for their insightful comments through which the quality of this work has been enhanced. The authors would also like to thank the associate editors and editor in chief for identifying the potential in this survey and taking it through the review process.

\bibliographystyle{ACM-Reference-Format}
\bibliography{citations}

\end{document}